\documentclass{article}

\usepackage{amssymb}
\usepackage{amsmath}
\usepackage{graphicx}
\usepackage{xcolor} 

\usepackage{caption}
\usepackage{tabularray}
\usepackage{varwidth}
\UseTblrLibrary{booktabs}
\usepackage{enumitem}
\usepackage{subcaption}
\usepackage{multirow}
\usepackage{microtype}
\usepackage{etoolbox} 

\usepackage[preprint]{corl_2025} 

\usepackage[capitalize]{cleveref}
\crefname{section}{Sec.}{Secs.}
\Crefname{section}{Section}{Sections}
\crefname{figure}{Fig.}{Figs.}
\Crefname{figure}{Figure}{Figures}
\crefname{table}{Tab.}{Tabs.}
\Crefname{table}{Table}{Tables}
\crefname{algorithm}{Algo.}{Algos.}
\Crefname{algorithm}{Algorithm}{Algorithms}

\newcommand{\argus}[1]{\textcolor{blue}{[MA] #1}}

\newcommand{\cm}{\checkmark}

\title{cVLA: Towards Efficient Camera-Space VLAs}

\author{
  \textbf{Max Argus},
  \textbf{Jelena Bratuli\'c},
  \textbf{Houman Masnavi},
  \textbf{Maxim Velikanov} \\ 
  \textbf{Nick Heppert},
  \textbf{Abhinav Valada}, 
  \textbf{Thomas Brox} \\
  University of Freiburg, Germany \\
  \texttt{argusm@cs.uni-freiburg.de}
}

\begin{document}
\maketitle


\begin{abstract}
Vision-Language-Action (VLA) models offer a compelling framework for tackling complex robotic manipulation tasks, but they are often expensive to train. In this paper, we propose a novel VLA approach that leverages the competitive performance of Vision Language Models (VLMs) on 2D images to directly infer robot end-effector poses in image frame coordinates.  Unlike prior VLA models that output low-level controls, our model predicts trajectory waypoints, making it both more efficient to train and robot embodiment agnostic. 
Despite its lightweight design, our next-token prediction architecture effectively learns meaningful and executable robot trajectories. 
We further explore the underutilized potential of incorporating depth images, inference-time techniques such as decoding strategies, and demonstration-conditioned action generation. Our model is trained on a simulated dataset and exhibits strong sim-to-real transfer capabilities. We evaluate our approach using a combination of simulated and real data, demonstrating its effectiveness on a real robotic system.
\end{abstract}

\keywords{VLAs,
    Manipulation, 
    Imitation Learning
}


\section{Introduction}

Vision-language-action (VLA) models integrate visual understanding with actionable decision-making by jointly learning from visual, linguistic, and interaction data. These methods enable fine-grained perception and action generation, allowing them to solve a diverse range of tasks~\cite{Driess2023PaLMEAE, Team2024OctoAO, Kim2024OpenVLAAO}. When trained on large-scale datasets of robot demonstrations, VLAs can generalize across a variety of robots and environments~\cite{Black20240AV}.
However, for further advancement of VLAs, we identified several key constraints:
a) Computational costs: Training VLAs demands significant computational resources, making experimentation challenging. 
b) Data limitations: Collecting high-quality, multimodal real-world datasets that pair all three modalities —vision, language, and interaction data —is expensive and time-consuming.
c) Evaluation and benchmarking: Standardized benchmarks for assessing VLAs performance often rely on real-world rollouts, making consistent comparisons difficult.
This work addresses these constraints by proposing a lightweight VLA system trained on a controllable synthetic dataset and designed for broad applicability across different domains.

Evaluating and potentially training in simulation may be a key method to address these problems. Its use is already widespread in other robotic areas, such as learning navigation and locomotion~\cite{Mirowski2018LearningTN, Savva2019HabitatAP, Lee2020LearningQL}, but this is not yet the case for VLAs. This may be due to the high precision required for control, the large number of degrees of freedom, and complex contact interactions~\cite{Mandlekar2018ROBOTURKAC, Batra2020RearrangementAC} and scene complexity. Bridging this gap remains a key research area in robotic manipulation. 
To this end, we utilize training from simulations by constructing a curated dataset with a strong camera viewpoint and object variation. The simulations and augmentations are carefully constructed to enable sim-to-real transfer.
Based on this data, we train a VLA based on the PaliGemma architecture~\cite{beyer2024paligemma}. We formulate our learning tasks as a one-step\footnote{Note that the description as 1-step models is sometimes also used to refer to models having only a single observation time-step as input, instead of the history~\cite{Li2024TowardsGR}.}
prediction of end-effector keyposes, which allows efficient training on numerous scenes. We evaluate our VLA systems on the DROID dataset \cite{Khazatsky2024DROIDAL}, simulations through ManiSkill \cite{Tao2024ManiSkill3GP}, and a real robot set-up.
Additionally, we explore the effect of defining keyposes in image frame coordinates instead of 6D end-effector poses.
Given our architecture's similarity to standard VLMs, we investigate several inference-time strategies, such as input image cropping and multiple prediction generation, and evaluate their impact on the final model performance. 

In this paper, rather than training a general-purpose foundation model, we focus on a narrow data distribution with a limited set of tabletop tasks, restricting ourselves to quasi-static manipulation and generating actions with low temporal resolution. We hope that our small-scale experiments can provide helpful insights into the factors affecting VLA performance and contribute to eventually scaling VLA systems. 
 Our contributions can be summarized as: 
 \begin{itemize}[topsep=-3px, noitemsep, leftmargin=15px]
     \item An efficient setup for training and evaluating 
     VLA models, with a diverse curated collection of shapes and texts, including a lightweight 1-shot imitation system.
     \item An investigation into inference time prediction strategies for VLAs and their evaluation, including a new decoding algorithm called beam-search-NMS.
     \item Public release of code, datasets, and models at \texttt{available upon acceptance}.
 \end{itemize}

\section{Related work}

\textbf{Vision-Language-Action Systems}
Recent VLA models integrate visual perception, language understanding, and action generation to achieve generalist robotic skills~\cite{Driess2023PaLMEAE, Team2024OctoAO, Kim2024OpenVLAAO}. A0~\citep{xu2025a0} introduces affordance-aware representations for cross-platform manipulation. TraceVLA~\citep{zheng2024tracevla} enhances spatial-temporal reasoning through visual trace prompts. GR00T N1~\citep{nvidia2025gr00tn1} scales VLA systems to humanoid robots by employing a dual deliberative and reactive system design, achieving strong generalization across different embodiments. Together, these works highlight progress toward unified high-level understanding and low-level control. 
Molmo~\cite{Deitke2024MolmoAP} and RoboPoint~\cite{Yuan2024RoboPointAV} take an intuitive but different view on the problem by introducing the concept of pointing directly in image space.

\textbf{Trajectory Prediction and Waypoint Representations} 
are critical for robust robot control. Inferred keyposes have been successfully applied to solve complex robotic manipulation tasks~\cite{shridhar2023perceiver}. Extending the concept, HDP~\cite{ma2024hierarchical} suggested connecting keyposes through diffusing low-level control actions. Most recently, PPI~\citep{yang2025gripper} introduces hybrid 6-DoF keyposes and pointflows to maintain spatial precision while supporting flexible closed-loop control, enabling superior two-arm manipulation. Such mid-level waypoint structures blend discrete and continuous cues, improving planning and execution across complex tasks.

\textbf{Training from Simulation} for VLA models enables scalable learning but faces challenges in sim-to-real transfer. DexGraspVLA~\citep{zhong2025dexgraspvla} combines a pre-trained vision-language planner with a diffusion-based controller, using a mix of real and simulated data to achieve robust zero-shot dexterous grasping. Robot manipulation benchmarks like CALVIN~\citep{mees2022calvin} and RLBench~\cite{james2019rlbench} provide simulated tasks to support large-scale model training. Advances in simulation domain randomization, heterogeneous datasets, and real-world alignment are key to bridging the sim-to-real gap. 

\textbf{Auxiliary Visual Tasks} help VLAs ground their predictions spatially. LLARVA~\citep{niu2024llarva} uses 2D trace supervision to align vision and action, improving task success rates. Gemini Robotics-ER~\citep{geminiroboticsteam2025geminiroboticsbringingai} leverages auxiliary outputs such as keypoint detection and motion sketching to enhance multistep reasoning and manipulation. Incorporating segmentation, depth estimation, and affordance prediction further improves generalization to unseen scenarios. 3D-VLA~\cite{Zhen20243DVLAA3} defines multiple auxiliary tasks and also takes depth values as inputs.

\textbf{Evaluation of VLAs} is investigated in a number of recent works. This includes evaluating VLAs trained primarily on real data through the use of aligned simulations called real-to-sim evaluations~\cite{Li2024EvaluatingRR, Wang2024TowardsTA}.
Long-horizon trajectory prediction is performed in VLA models for autonomous driving~\cite{Arai2024CoVLACV}, and in some cases, also considering metrics sensitive to the diversity of predictions~\cite{Chen2023CRITERIAAN}.

\textbf{Few-Shot Imitation from Demonstrations} remains a key challenge in robotics, particularly when scaling efficiently to new tasks. Early approaches introduced meta-learning for one-shot imitation using task-specific demonstration-action pairs~\citep{duan2017oneshot} and employing metric learning to embed demonstrations, enabling strong retrieval-based generalization~\citep{james2018task}. Recent works explore data-efficient retrieval by selectively utilizing extensive unlabeled datasets~\cite{du2023behavior} and enhancing retrieval with optical flow representations~\citep{lin2024flowretrieval}. One-shot methods~\citep{Zhang2024Invariance, heppert2024ditto} achieve success with a single unannotated demonstration, while \citet{dipalo2024kat} demonstrate a few-shot visual imitation through in-context learning with pre-trained transformers. Parallel to us, \citet{fu2024context} proposes an in-context imitation learning method which requires training and context data collected in the same environment.

\section{Technical Approach}

\begin{figure}[t]
    \centering
    \includegraphics[width=\linewidth]{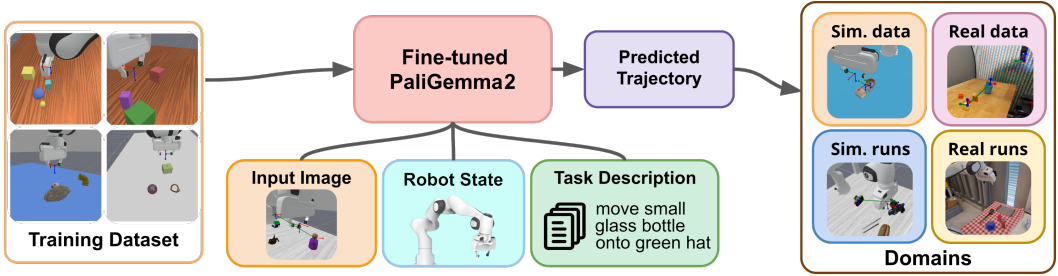}
    \caption{\textbf{Overview of cVLA.} Our lightweight method is based on fine-tuning a PaliGemma2~\cite{Steiner2024PaliGemma2A} model for trajectory prediction using our curated dataset with a single image, robot state, and task description as inputs. Our synthetic training dataset is built from different simulations of pick-and-place tasks, which enables easy scaling and an efficient training pipeline. The approach shows good generalization across different application domains, including simulation, real data, and real robot setups, and offers a simpler setup for experimental research and development of VLAs.\looseness=-1}
    \label{fig:method_figure}
\end{figure}

In the following, we detail our technical approach, summarized in \cref{fig:method_figure}. First, we describe our base model architecture in \cref{subsec:approach:base_model}. Then, we introduce our novel action representations, combining image-frame coordinates and camera frame-poses in \cref{subsec:approach:action_reprs}. Thirdly, we explain how depth information is incorporated into our model in \cref{subsec:approach:depth}. Finally, \cref{subsec:approach:imitation} outlines the extension of the base method to few-shot trajectory imitation set-up.

\subsection{Base Model}
\label{subsec:approach:base_model}
We fine-tune a pre-trained vision language model (VLM), in our case PaliGemma2~\cite{Steiner2024PaliGemma2A}. By using an already pre-trained model, the result is an efficient and robust VLA system. Following standard VLA prompting conventions, we design our prompts as:
    \texttt{<live img(s)>} + \texttt{<robot state>} + \texttt{<task description>} $\rightarrow$ \texttt{<estimated trajectory>},
where \texttt{<live img(s)>} are an RGB and optional depth image (see \cref{subsec:approach:depth}), \texttt{<robot state>} is the current end-effector pose of the robot and \texttt{<task description>} an instruction in natural language. For \texttt{<estimated trajectory>}, unlike most other VLAs, we also make the following design decisions: a) instead of predicting a full trajectory, we predict trajectory keyposes, of which we predict only two. These are then converted into a robot trajectory using a low-level planner. Then b) we also make a one-step prediction, i.e., we predict the entire trajectory for a scene in one step. This choice has the drawback of restricting the flexibility of the system, however it has the advantage of making training more efficient.\footnote{Again here we do not aim to train a foundation model, however strategies like increasing temporal resolution is a possible extension. 
}

Our efficient VLA model is fine-tuned only on attention layer parameters, which, although a simple and lightweight modification, ensures creation of a strong trajectory prediction model for our use cases, which also allows investigating inference-time strategies. The training procedure and hyperparameters are described in detail in the appendix.

\textbf{Action Representations}: \label{subsec:approach:action_reprs}
Similar to RT-1~\cite{Brohan2022RT1RT}, we encode 6-DoF gripper poses using discretized tokens. PaliGemma2~\cite{Steiner2024PaliGemma2A} contains special tokens for image detection and segmentation, which we repurpose for pose prediction similar to \cite{Ning2023AllIT}. Instead of encoding actions as end-effector deltas in robot frame coordinates, we encode actions as absolute positions in either the robot-base coordinates or image-frame coordinates, i.e., the normalized width, height, and distance from the camera. We extend the localization tokens (n=1024) to also predict depth, given the gripper position. Orientations are encoded using the segmentation tokens~(n=128). The same scheme is used for both image-frame and robot-frame actions. Additionally, we experiment with using a smaller number of tokens to predict the position (n=512, 256, 128). This frees up the tokens to be used for predicting depth separately, an approach we also evaluate. An example of the logit distribution is shown in \cref{fig:beam_search_example}.

\textbf{Depth Input}:\label{subsec:approach:depth}
We extend our approach to utilize depth observations as input. To leverage the strong image encoders available for RGB images, we convert our depth maps into RGB using Matplotlib's viridis color map. These images are then processed with the same pre-trained image encoder as the natural images.

\subsection{Robotic Imitation}
\label{subsec:approach:imitation}
We extend our approach to few-shot imitation learning by  conditioning trajectory prediction on demonstration image-trajectory pairs instead of a natural language text description. The system infers the task from the given demonstration image-trajectory pair and must apply it to a new scene image - similar to in-context imitation learning, but in our case, we train the model to learn how to do imitation. We do not perform any fine-tuning on the novel scene image during inference time. 

Our extended approach now introduces a multi-step reasoning process: given the context template of \texttt{<demo img>} + \texttt{<demo trajectory>} + \texttt{<live img>} $\rightarrow$ \texttt{<estimated trajectory>}, the model must infer the task from the demonstration image (input) and trajectory (output) pair, map the object positions to the associated tokens, and establish the correspondence between the objects in the new scene and the predicted trajectory. At test time, we sample demonstration pairs from hold-out data where the task is shared between the demonstration pair and the live image. 

To enable this, we expand our training datasets by building a task-demonstration sampler. We build a look-up table of available tasks and generate a large number of random demonstration-query pairs from available scenes. Every scene can be seen only once in the pair, either as a demonstration or as a query. Due to the high number of possible demonstration pairs, we fine-tune the model for 16k iterations, keeping other hyperparameters unchanged from the original setting. Further details and exact prompt examples are provided in the appendix.

\section{Dataset Generation}

Our VLA sysem makes use of simulated data for training, and a combination of simulated and real data for evaluation. In the following, we describe the data collection procedure.

\subsection{Simulated Training Dataset Generation}
We use the ManiSkill~\cite{Tao2024ManiSkill3GP} simulator to create our environments.  The following outlines our data generation procedure, the 3D objects used, and our suggested augmentation strategies.

\textbf{Generation Procedure}: 
To generate a new data sample, we spawn a set of objects and an instruction, and then calculate the target object pose.
We then use an analytical grasp model to generate grasps on the object and use the privileged information of known object poses to calculate a release pose. 
While this step can be easily performed offline to speed up generation time, we also extend our simulation to execute the task and actually verify task success. For further information, see \cref{app:sec:simulation}.

\textbf{3D Model Assets}: 
We use two different sets of object assets -- a set of simple geometric shapes and a set of real-world objects, scraped from the Objaverse dataset~\citep{deitke2023objaverse}. 
We show example images of the objects in \cref{app:sec:datasets}.

\textit{\textbf{CLEVR}}: Simple geometric shapes of different sizes, inspired by the CLEVR~\cite{Johnson2016CLEVRAD} dataset. Specifically, our environment consisted of three shapes: a cube, a block, and a sphere; two different sizes, with diameters of 7 cm and 3.5 cm, respectively; and eight colors.

\textit{\textbf{Objaverse}}: 
To create a more diverse training and testing scenario, we construct a simulation with a large and diverse set of assets. To automatically generate text instructions for training, we require realistically scaled models combined with a concise text description. This is done similarly to Spoc~\cite{ehsani2024spoc}, using the Objaverse-1M~\cite{deitke2023objaverse} dataset and described in detail in \cref{sec:objaverse_curation}.

\textbf{Augmentations and Randomization}:
We apply standard image augmentations, similar to \cite{ehsani2024spoc}.
Additionally, we perform background image randomization, similar to Kubric
~\cite{Greff2022KubricAS}, using indoor scene images from \cite{Quattoni2009RecognizingIS} to replace everything except the robot and objects, with a probability of 0.2. We perform text randomization by constructing text templates from the training split of our test data and filling in the relevant object names.

We also include randomization in the scene generation pipeline, offering easier and harder versions of the dataset. The easier version features less randomization, while the harder version includes severe scene and camera field-of-view randomization. For the harder variant of the dataset, we additionally perform a visibility test to ensure that only physically plausible environments are considered. Thus, we have four variants of the training data: CLEVR-easy, CLEVR-hard, Mix-easy, and Mix-hard. Simulation experiments are conducted using either the CLEVR variants or the Objaverse-easy and Objaverse-hard sets.

\subsection{Real Evaluation Data Generation}

We use sequences from DROID~\cite{Khazatsky2024DROIDAL} dataset, an existing robot manipulation dataset, to evaluate our model's performance. It has a diverse mix of scenes and actions, as well as the extrinsic calibration information necessary to evaluate our system. However, the quality of the extrinsic calibration is inconsistent, thus we need to manually filter the data using the projection of the end-effector position into the image. For further information, see \cref{app:sec:datasets}.

To simplify the evaluation, and since our training data only involves move-A-to-B actions, we extract two subsets from DROID~\cite{Khazatsky2024DROIDAL} in which cubes are moved in this manner. From these sequences, we take only the initial frames. The first subset, \textit{\textbf{DROID-hard}}, includes images with confounding objects. It is created to test the model's ability to predict the multi-modal distribution of trajectories. The second, \textit{\textbf{DROID-easy}}, has confounding objects blurred out, creating an easier setting in which to test generalization performance.

For offline evaluation, we calculate the L1 error for positions and rotations between predicted and ground-truth poses. For further details see \cref{app:sec:lonemetric}.
\section{Experiments}

We evaluate several aspects of our VLA system. First, the effect of various design choices on performance within the simulation domain in \cref{sec:experiments_ablation}. Second, how our model can be used to do one-shot imitation in \cref{sec:experiments_imitation}. Third, we investigate how inference time strategies can be used to boost performance in \cref{sec:experiments_inference}. 
Finally, we show zero-shot inference on a real robot without any real-world fine-tuning in \cref{sec:experiments:real_robot}.

\subsection{Action Encoding, Depth, and Domain Randomization Ablations}
\label{sec:experiments_ablation}
As described earlier, our method leverages two different versions of the training dataset, followed by visual and textual enhancements. In \cref{tab:ablation}, we evaluate the influence of each component on the final model as well as including auxiliary depth information in simulation (see \cref{app:sec:simulation} for further information about the simulation setup). 
All methods are trained on a harder version of the mix dataset, which includes camera and scene randomization.

\begin{table}[t]
    \footnotesize
    \centering
    \caption{\textbf{Ablation study on simulation success rate.} We evaluate using CLEVR or Objaverse assets for differently randomized versions of the simulation (where easy uses fewer cameras and scene randomness). We observe clear patterns; adding depth to the prompt improves performance in all scenarios, and training with augmentation harms simulation performance.
    }
    \label{tab:ablation}
    \begin{tabular}{cccc|cc|cc}
        \toprule
            \multirow{2}{*}{CLEVR} & 
            \multirow{2}{*}{Objaverse} & 
            \multirow{2}{*}{Augs.} & 
            \multirow{2}{*}{Depth} & 
            \multicolumn{2}{c|}{Objaverse SR~($\uparrow)$} & 
            \multicolumn{2}{c}{CLEVR SR~($\uparrow)$} \\
            & & & & easy & hard  & easy  & hard \\
        \midrule
        \cm & & &             & 8\%    &  4\%  & 40\%  &  42\%           \\
        \cm && \cm  &         & 0\%    &  4\%  & 46\%  &  26\%           \\
        \cm & \cm  &&         & 18\% & 24\%    & 44\%  &  42\%           \\
        \cm & \cm  & \cm &    & 18\% & 20\%    & 34\%  & 32\%            \\
        \cm & & \cm & \cm     & 6\%    &  6\%  & \textbf{56\%}  & 50\%        \\
        \cm & \cm & \cm & \cm & \textbf{20\%} & \textbf{30\%}    & 52\%  & \textbf{54\%}             \\
        \bottomrule
    \end{tabular}
    \vspace{-1em}
\end{table}

We observe that adding depth information to the model significantly improves performance in simulation success rates and results in fewer drastic failure cases. Moreover, training solely on CLEVR assets improves performance on CLEVR-based simulations, but fails to generalize to Objaverse-based simulations, demonstrating the need for diverse 3D assets. 

Next, we compare different action representation schemes in  \cref{fig:action_reps}. We compare the performance in terms of success rate on the CLEVR-easy simulation, where we observe that the camera frame performs better on average. This likely underestimates the utility of image frame coordinates, as in these simple environments, it is easier to overfit, e.g., on gripper appearance and camera intrinsics.

\subsection{One-Shot Imitation Experiments}
\label{sec:experiments_imitation}
We further evaluate our system for simple one-shot trajectory-conditioned imitation from demonstrations, where the task is inferred from a single demonstration consisting of an image and a trajectory, rather than a natural language description.
As described in \cref{subsec:approach:imitation}, this set-up poses additional challenges, since the task needs to be deduced from a multi-step reasoning chain.

\begin{table}[t]
    \footnotesize
    \centering
    \caption{\textbf{One-shot imitation with demonstrations.} Our efficient pipeline can be easily extended into an imitation learning model, which achieves good success rates in robotic simulations. 
    }
    \begin{tabular}{c|cc|ccc}
        \toprule
        Train& \multicolumn{2}{c|}{CLEVR Sim. SR~($\uparrow$)}  & \multicolumn{3}{c}{Static data traj. L1 error ($\downarrow$) }\\ 
        Data & easy          & hard         & DROID-easy & Simulation data & Objaverse-easy \\
        \midrule
        CLEVR-easy & \textbf{70 \%}           & 18\%   &  16.37        &  \textbf{3.19}               &   15.31           \\
        CLEVR-hard & 44\%            & \textbf{28\%}    &  \textbf{11.56}        &  6.41               &   \textbf{14.37}             \\ 
        \bottomrule
    \end{tabular}
    
    \label{tab:imitation_results}
\end{table}

We train the imitation model only on CLEVR versions of the dataset, including both CLEVR-easy (with less camera and scene randomization) and CLEVR-hard (with more camera and scene randomization), and report the success rate in simulation.
We further evaluate the performance on real data of the simplest variant of the DROID dataset, using hold-out validation data whose distributions are aligned with the training distributions of CLEVR-easy and CLEVR-hard. Finally, we evaluate generalization to novel objects in the scene on the Objaverse version of the data.

Results are show in \cref{tab:imitation_results}. We report a success rate of 70\% for the easy version of the dataset and 28\% on the harder setup. Furthermore, we observe better results on real data and generalization data for the dataset trained with a harder version of itself, showing that camera and scene randomization are essential for achieving robustness.
Visualizations of predicted trajectories and demonstration pairs are available in \cref{app:sec:one_shot_details}.

\subsection{Inference-Time Strategies}
\label{sec:experiments_inference}

In addition to the discussed technical and dataset advances, we also evaluate recent trends in VLMs and their impact on VLAs' performance, namely image cropping and next-token decoding strategies. 
\begin{figure}[ht]
    \centering

    \begin{minipage}[t]{0.48\textwidth}
        \centering
        \includegraphics[width=.95\linewidth]{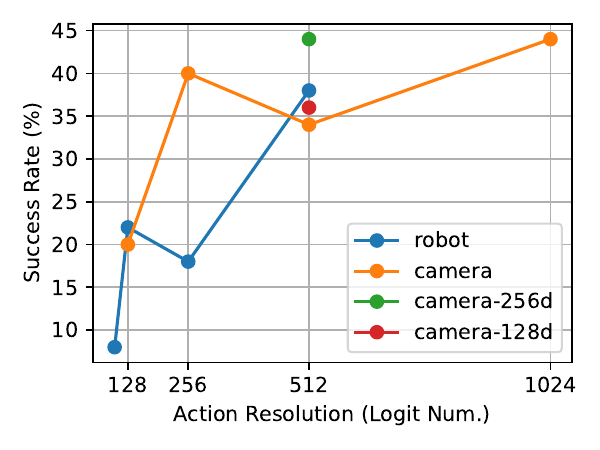}
        \captionof{figure}{\textbf{Action representation ablation.} Comparing robot and image coordinate frame action predictions, success rate on CLEVR-easy simulation, with camera frame performing better on average.}
        \label{fig:action_reps}
    \end{minipage}
    \hfill
    \begin{minipage}[t]{0.48\textwidth}
        \centering
        \includegraphics[trim=4 0 4 23pt, clip, width=.95\linewidth]{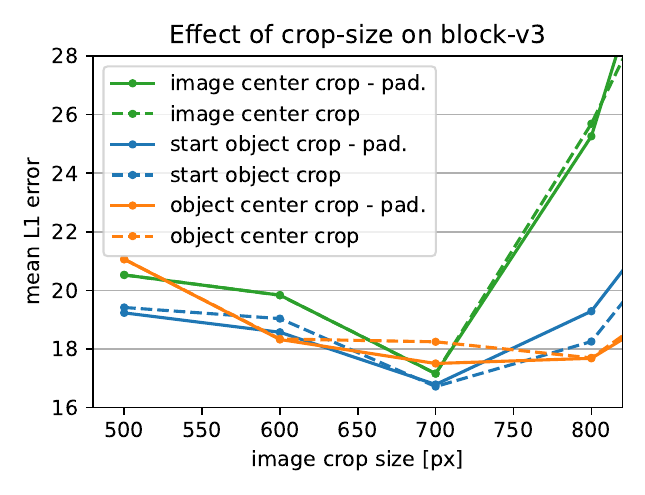}
        \captionof{figure}{\textbf{Cropping strategies comparison.} Cropping can consistently improve performance, but also starts inducing failures, on DROID-hard}
        \label{fig:effect_of_crop_size_block_v3}
    \end{minipage}

    \vspace{-1em}
\end{figure}


\textbf{Input Image Cropping.}
Accurate object localization is crucial for successful robotic manipulation, as even small errors in identifying object positions can result in task failure. In our system, two factors contribute to sensitivity in localization: (a) the model operates on relatively low-resolution input images \mbox{($224 \times 224$ px)}, and (b) we predict trajectories in a single step without iterative refinement. As a result, the model’s performance is susceptible to the scale of the objects within the image; smaller objects may not be resolved clearly enough to enable precise keyposes prediction. 
To address this, we investigate the impact of different image cropping strategies. See \cref{fig:effect_of_crop_size_block_v3} for the results and \cref{sec:cropping_details} for details. Cropping significantly improves performance and is used in subsequent experiments.


\textbf{Multiple Prediction Generation and Evaluation.}
In language generation, decoding strategies approximate the most probable token sequence under the model. While greedy decoding is the default, alternative methods can improve quality by identifying high-probability sequences, but at the cost of increased computation. Generating multiple plausible predictions also enables evaluation of both solution accuracy and ability to make diverse predictions.
We tested the following standard decoding approaches: 
Beam search, which keeps $n$ most probable candidate sequences at each decoding step, and sampling, which diversifies the output by sampling from the token probability distribution.

\begin{figure}[t]
    \centering
    \begin{subfigure}[t]{.48\textwidth}
      \centering
       \includegraphics[width=0.65\columnwidth
       ]{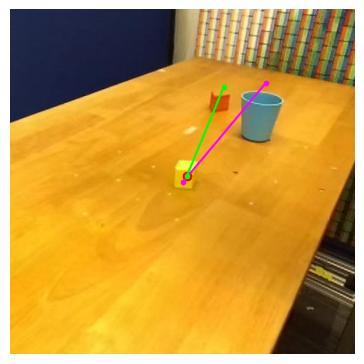}
      \caption{Input image with task prompt \texttt{"put the yellow block in the cup"}.}
    \end{subfigure}\hfill
    \begin{subfigure}[t]{.48\textwidth}
      \centering
        \includegraphics[width=1.0\columnwidth
        ]{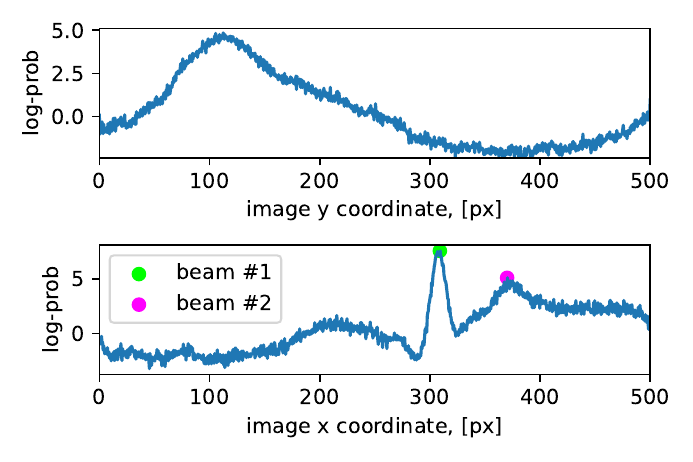}  
        \caption{Logit distribution of the x and y position tokens of the target location.}
    \end{subfigure}
    \caption{
        \textbf{Exemplary motivation for decoding.} 
        We qualitatively visualize results on episode 81 of the DROID-hard dataset. The most probable beam corresponds to the red cube, but our proposed NMS-based beam decoding strategy also detects the correct target object location (blue cup).
    }
    \label{fig:beam_search_example}
\end{figure}

Our contribution is a custom decoding strategy for VLA models, {\bf beam-search-NMS} (Non-Maximum Suppression). We observe that the predicted distribution of VLA models for dense pose tokens behaves differently from sparse language tokens (\cref{fig:beam_search_example}), they are smooth and have multiple peaks, so choosing the top k tokens often results in almost the same pose. To find the peaks of the distribution, we do beam search ($n=3$) and search for local maxima with non-maximum suppression within a window of size $w=100$, see \cref{sec:app-beam-search-NMS} for details and \cref{tab:trajectory_errors} for results.

To evaluate the distribution of predicted trajectories, we propose using mean Average Precision (mAP)—either with respect to success rate (SR) in simulation as is done in \cite{Fang2022AnyGraspRA, Zhou2024VariationalDO} or thresholded Euclidean distance in offline settings. This metric offers a more informative assessment of distributional accuracy than traditional pointwise comparisons such as L1 or L2 distances.
For manipulation, we suggest $\text{mAP}_{[0.5\,50]}$, meaning the mAP over L1 distances APs at thresholds of [.5, 1, 2, 5, 10, 20, 50]~cm, with 1cm = 10 degrees. AP calculation and curves are shown in \cref{app:sec:additional_experimental}.

\begin{table}[t]
    \footnotesize
    \centering
    \caption{
        \textbf{Results for different decoding strategies.} For all methods, we select exactly one prediction for each episode and compute the mean L1 error between the ground truth and predicted trajectories. All methods except greedy are combined with beam search with $n=3$ beams. The top-3 row shows the lowest error among the predicted beams. 
    }
    \begin{tabular}{l|ccccc}
        \toprule
        & Greedy  search & Sampling & Beam search  & Beam search-NMS  \\
        \midrule
        Top-1 & 34.44 & 34.31 &  34.17 & \textbf{33.42} \\
        Top-3 & - & 33.94 & 33.94 & \textbf{25.00} \\
        \bottomrule
    \end{tabular}
    \label{tab:trajectory_errors}
    \vspace{2em}
\end{table}

\subsection{Real Robot Setup and Experiments}
\label{sec:experiments:real_robot}
We conduct our experiments using a Franka Panda robotic arm mounted on a mobile base, performing 15 distinct tabletop manipulation tasks involving everyday household items. Our setup supports both external-view cameras and wrist-mounted vision systems. For convenience, we use the wrist-mounted camera, the StereoLabs ZED2i. To ensure our method is robust to different viewpoints, we randomize the robots starting position in each trial. \cref{fig:real_world_experiment} presents two sample scenarios.
Example videos are included in the supplementary.

\begin{figure}[t]
    \centering
    \includegraphics[width=1.0\linewidth]{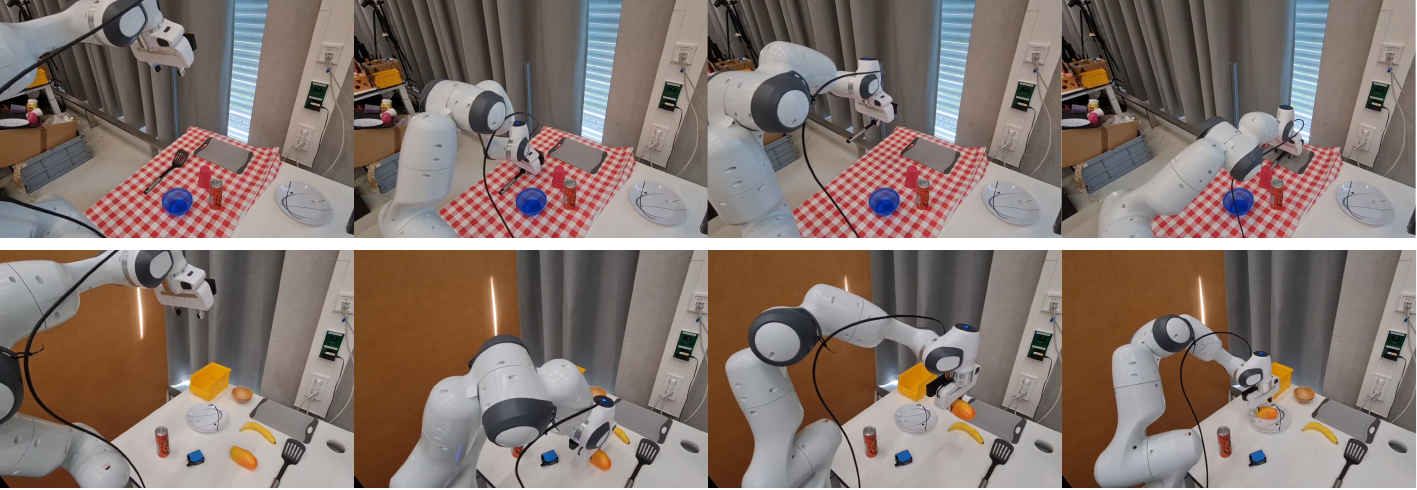}
    \caption{\textbf{Real-world demonstration of our approach.} The top row illustrates the task of placing a spatula onto a cutting board, while the bottom row depicts the robot placing a mango onto a plate.}
    \label{fig:real_world_experiment}
\end{figure}

\section{Conclusion}

We present an efficient VLA that is trained using image frame coordinates and makes a direct 1-step prediction of two end-effector keyposes. While this system is limited in flexibility and is not a foundation model, it is well-suited to run a wide range of VLA experiments, including ablations, 1-shot imitation, and the applicability of LLM inference strategies to VLA tasks. Finally, we show that our system is applicable in realrobot experiments without any fine-tuning. We believe that it can provide a good foundation for further research into simulation-trained VLMs.

\section{Limitations}

While our work demonstrates the effectiveness of a lightweight VLA system for keypose prediction in image frame coordinates, it has several limitations that constrain its applicability and generalizability. First, the model is evaluated on a single manipulation task involving small objects and top-down grasps. As such, the learned policy may not transfer well to more diverse tasks, larger objects, or more complex grasping strategies (e.g., side grasps or in-hand manipulation). Second, although we include rotation information in keypose predictions, the system exhibits poor rotation accuracy on real-world data, which limits its effectiveness in tasks that require precise orientation control. Finally, the model is trained and evaluated in simulation with limited real-world testing.
This suggests a need for future work on improving robustness and generalization,  potentially better data generation. Additionally, broader testing across embodiments
and task types would be necessary to establish the scalability and reliability of the proposed system.


	


\acknowledgments{This research was funded by the German Research Foundation (DFG) 417962828, 539134284, and 499552394 (SFB 1597 - Small Data). This work was partially funded by the Carl Zeiss Foundation with the ReScaLe project. Nick Heppert is supported by the Konrad Zuse School of Excellence in Learning and Intelligent Systems (ELIZA) through the DAAD programme Konrad Zuse Schools of Excellence in Artificial Intelligence, sponsored by the Federal Ministry of Education and Research.}


\bibliography{main}  

\begin{thebibliography}{51}
\providecommand{\natexlab}[1]{#1}
\providecommand{\url}[1]{\texttt{#1}}
\expandafter\ifx\csname urlstyle\endcsname\relax
  \providecommand{\doi}[1]{doi: #1}\else
  \providecommand{\doi}{doi: \begingroup \urlstyle{rm}\Url}\fi

\bibitem[Driess et~al.(2023)Driess, Xia, Sajjadi, Lynch, Chowdhery, Ichter, Wahid, Tompson, Vuong, Yu, Huang, Chebotar, Sermanet, Duckworth, Levine, Vanhoucke, Hausman, Toussaint, Greff, Zeng, Mordatch, and Florence]{Driess2023PaLMEAE}
D.~Driess, F.~Xia, M.~S.~M. Sajjadi, C.~Lynch, A.~Chowdhery, B.~Ichter, A.~Wahid, J.~Tompson, Q.~H. Vuong, T.~Yu, W.~Huang, Y.~Chebotar, P.~Sermanet, D.~Duckworth, S.~Levine, V.~Vanhoucke, K.~Hausman, M.~Toussaint, K.~Greff, A.~Zeng, I.~Mordatch, and P.~R. Florence.
\newblock Palm-e: An embodied multimodal language model.
\newblock In \emph{International Conference on Machine Learning}, 2023.

\bibitem[Ghosh et~al.(2024)Ghosh, Walke, Pertsch, Black, Mees, Dasari, Hejna, Kreiman, Xu, Luo, Tan, Sanketi, Vuong, Xiao, Sadigh, Finn, and Levine]{Team2024OctoAO}
D.~Ghosh, H.~R. Walke, K.~Pertsch, K.~Black, O.~Mees, S.~Dasari, J.~Hejna, T.~Kreiman, C.~Xu, J.~Luo, Y.~L. Tan, P.~R. Sanketi, Q.~Vuong, T.~Xiao, D.~Sadigh, C.~Finn, and S.~Levine.
\newblock Octo: An open-source generalist robot policy.
\newblock \emph{arXiv preprint arXiv:2405.12213}, 2024.

\bibitem[Kim et~al.(2024)Kim, Pertsch, Karamcheti, Xiao, Balakrishna, Nair, Rafailov, Foster, Lam, Sanketi, Vuong, Kollar, Burchfiel, Tedrake, Sadigh, Levine, Liang, and Finn]{Kim2024OpenVLAAO}
M.~J. Kim, K.~Pertsch, S.~Karamcheti, T.~Xiao, A.~Balakrishna, S.~Nair, R.~Rafailov, E.~Foster, G.~Lam, P.~R. Sanketi, Q.~Vuong, T.~Kollar, B.~Burchfiel, R.~Tedrake, D.~Sadigh, S.~Levine, P.~Liang, and C.~Finn.
\newblock Openvla: An open-source vision-language-action model.
\newblock \emph{arXiv preprint arXiv:2406.09246}, 2024.

\bibitem[Black et~al.(2024)Black, Brown, Driess, Esmail, Equi, Finn, Fusai, Groom, Hausman, Ichter, Jakubczak, Jones, Ke, Levine, Li-Bell, Mothukuri, Nair, Pertsch, Shi, Tanner, Vuong, Walling, Wang, and Zhilinsky]{Black20240AV}
K.~Black, N.~Brown, D.~Driess, A.~Esmail, M.~Equi, C.~Finn, N.~Fusai, L.~Groom, K.~Hausman, B.~Ichter, S.~Jakubczak, T.~Jones, L.~Ke, S.~Levine, A.~Li-Bell, M.~Mothukuri, S.~Nair, K.~Pertsch, L.~X. Shi, J.~Tanner, Q.~Vuong, A.~Walling, H.~Wang, and U.~Zhilinsky.
\newblock $\pi$0: A vision-language-action flow model for general robot control.
\newblock \emph{arXiv preprint arXiv:2410.24164}, 2024.

\bibitem[Mirowski et~al.(2018)Mirowski, Grimes, Malinowski, Hermann, Anderson, Teplyashin, Simonyan, Kavukcuoglu, Zisserman, and Hadsell]{Mirowski2018LearningTN}
P.~W. Mirowski, M.~K. Grimes, M.~Malinowski, K.~M. Hermann, K.~Anderson, D.~Teplyashin, K.~Simonyan, K.~Kavukcuoglu, A.~Zisserman, and R.~Hadsell.
\newblock Learning to navigate in cities without a map.
\newblock \emph{arXiv preprint arXiv:1804.00168}, 2018.

\bibitem[Savva et~al.(2019)Savva, Kadian, Maksymets, Zhao, Wijmans, Jain, Straub, Liu, Koltun, Malik, Parikh, and Batra]{Savva2019HabitatAP}
M.~Savva, A.~Kadian, O.~Maksymets, Y.~Zhao, E.~Wijmans, B.~Jain, J.~Straub, J.~Liu, V.~Koltun, J.~Malik, D.~Parikh, and D.~Batra.
\newblock Habitat: A platform for embodied ai research.
\newblock \emph{Conference on Computer Vision (ICCV)}, pages 9338--9346, 2019.

\bibitem[Lee et~al.(2020)Lee, Hwangbo, Wellhausen, Koltun, and Hutter]{Lee2020LearningQL}
J.~Lee, J.~Hwangbo, L.~Wellhausen, V.~Koltun, and M.~Hutter.
\newblock Learning quadrupedal locomotion over challenging terrain.
\newblock \emph{Science Robotics}, 5, 2020.

\bibitem[Mandlekar et~al.(2018)Mandlekar, Zhu, Garg, Booher, Spero, Tung, Gao, Emmons, Gupta, Orbay, Savarese, and Fei-Fei]{Mandlekar2018ROBOTURKAC}
A.~Mandlekar, Y.~Zhu, A.~Garg, J.~Booher, M.~Spero, A.~Tung, J.~Gao, J.~Emmons, A.~Gupta, E.~Orbay, S.~Savarese, and L.~Fei-Fei.
\newblock Roboturk: A crowdsourcing platform for robotic skill learning through imitation.
\newblock In \emph{Conference on Robot Learning (CoRL)}, 2018.

\bibitem[Batra et~al.(2020)Batra, Chang, Chernova, Davison, Deng, Koltun, Levine, Malik, Mordatch, Mottaghi, Savva, and Su]{Batra2020RearrangementAC}
D.~Batra, A.~X. Chang, S.~Chernova, A.~J. Davison, J.~Deng, V.~Koltun, S.~Levine, J.~Malik, I.~Mordatch, R.~Mottaghi, M.~Savva, and H.~Su.
\newblock Rearrangement: A challenge for embodied ai.
\newblock \emph{arXiv preprint arXiv:2011.01975}, 2020.

\bibitem[Beyer et~al.(2024)Beyer, Steiner, Pinto, Kolesnikov, Wang, Salz, Neumann, Alabdulmohsin, Tschannen, Bugliarello, et~al.]{beyer2024paligemma}
L.~Beyer, A.~Steiner, A.~S. Pinto, A.~Kolesnikov, X.~Wang, D.~Salz, M.~Neumann, I.~Alabdulmohsin, M.~Tschannen, E.~Bugliarello, et~al.
\newblock Paligemma: A versatile 3b vlm for transfer.
\newblock \emph{arXiv preprint arXiv:2407.07726}, 2024.

\bibitem[Li et~al.(2024)Li, Li, Liu, Wang, Liu, Kang, Ma, Kong, Zhang, and Liu]{Li2024TowardsGR}
X.~Li, P.~Li, M.~Liu, D.~Wang, J.~Liu, B.~Kang, X.~Ma, T.~Kong, H.~Zhang, and H.~Liu.
\newblock Towards generalist robot policies: What matters in building vision-language-action models.
\newblock \emph{arXiv preprint arXiv:2412.14058}, 2024.

\bibitem[Khazatsky et~al.(2024)Khazatsky, Pertsch, Nair, Balakrishna, Dasari, Karamcheti, Nasiriany, Srirama, Chen, Ellis, Fagan, Hejna, Itkina, Lepert, Ma, Miller, Wu, Belkhale, Dass, Ha, Jain, Lee, Lee, Memmel, Park, Radosavovic, Wang, Zhan, Black, Chi, Hatch, Lin, Lu, Mercat, Rehman, Sanketi, Sharma, Simpson, Vuong, Walke, Wulfe, Xiao, Yang, Yavary, Zhao, Agia, Baijal, Castro, Chen, Chen, Chung, Drake, Foster, Gao, Herrera, Heo, Hsu, Hu, Irshad, Jackson, Le, Li, Lin, Lin, Ma, Maddukuri, Mirchandani, Morton, Nguyen, O'Neill, Scalise, Seale, Son, Tian, Tran, Wang, Wu, Xie, Yang, Yin, Zhang, Bastani, Berseth, Bohg, Goldberg, Gupta, Gupta, Jayaraman, Lim, Malik, Mart'in-Mart'in, Ramamoorthy, Sadigh, Song, Wu, Yip, Zhu, Kollar, Levine, and Finn]{Khazatsky2024DROIDAL}
A.~Khazatsky, K.~Pertsch, S.~Nair, A.~Balakrishna, S.~Dasari, S.~Karamcheti, S.~Nasiriany, M.~K. Srirama, L.~Y. Chen, K.~Ellis, P.~Fagan, J.~Hejna, M.~Itkina, M.~Lepert, Y.~Ma, P.~T. Miller, J.~Wu, S.~Belkhale, S.~Dass, H.~Ha, A.~Jain, A.~Lee, Y.~Lee, M.~Memmel, S.~Y. Park, I.~Radosavovic, K.~Wang, A.~Zhan, K.~Black, C.~Chi, K.~B. Hatch, S.~Lin, J.~Lu, J.-P. Mercat, A.~Rehman, P.~R. Sanketi, A.~Sharma, C.~B. Simpson, Q.~U. Vuong, H.~R. Walke, B.~Wulfe, T.~Xiao, J.~H. Yang, A.~Yavary, T.~Zhao, C.~Agia, R.~Baijal, M.~G. Castro, D.~L. Chen, Q.~Chen, T.~Chung, J.~Drake, E.~P. Foster, J.~Gao, D.~A. Herrera, M.~Heo, K.~Hsu, J.~Hu, M.~Z. Irshad, D.~Jackson, C.~Le, Y.~Li, K.~Lin, R.~Lin, Z.~Ma, A.~Maddukuri, S.~Mirchandani, D.~Morton, T.~Nguyen, A.~O'Neill, R.~M. Scalise, D.~Seale, V.~Son, S.~Tian, E.~Tran, A.~Wang, Y.~Wu, A.~Xie, J.~Yang, P.~Yin, Y.~Zhang, O.~Bastani, G.~Berseth, J.~Bohg, K.~Goldberg, A.~Gupta, A.~Gupta, D.~Jayaraman, J.~J. Lim, J.~Malik, R.~Mart'in-Mart'in, S.~Ramamoorthy, D.~Sadigh, S.~Song,
  J.~Wu, M.~C. Yip, Y.~Zhu, T.~Kollar, S.~Levine, and C.~Finn.
\newblock Droid: A large-scale in-the-wild robot manipulation dataset.
\newblock \emph{arXiv preprint arXiv:2403.12945}, 2024.

\bibitem[Tao et~al.(2024)Tao, Xiang, Shukla, Qin, Hinrichsen, Yuan, Bao, Lin, Liu, kai Chan, Gao, Li, Mu, Xiao, Gurha, Huang, Calandra, Chen, Luo, and Su]{Tao2024ManiSkill3GP}
S.~Tao, F.~Xiang, A.~Shukla, Y.~Qin, X.~Hinrichsen, X.~Yuan, C.~Bao, X.~Lin, Y.~Liu, T.~kai Chan, Y.~Gao, X.~Li, T.~Mu, N.~Xiao, A.~Gurha, Z.~Huang, R.~Calandra, R.~Chen, S.~Luo, and H.~Su.
\newblock Maniskill3: Gpu parallelized robotics simulation and rendering for generalizable embodied ai.
\newblock \emph{arXiv preprint arXiv:2410.00425}, 2024.

\bibitem[Xu et~al.(2025)Xu, Zhang, Guo, Wen, Yang, Lin, Huang, Li, Zhang, Wang, Kuang, Cao, Zheng, and Liang]{xu2025a0}
R.~Xu, J.~Zhang, M.~Guo, Y.~Wen, H.~Yang, M.~Lin, J.~Huang, Z.~Li, K.~Zhang, L.~Wang, Y.~Kuang, M.~Cao, F.~Zheng, and X.~Liang.
\newblock A0: An affordance-aware hierarchical model for general robotic manipulation.
\newblock \emph{arXiv preprint arXiv:2504.12636}, 2025.

\bibitem[Zheng et~al.(2024)Zheng, Liang, Huang, Gao, Daum{\'e}~III, Kolobov, Huang, and Yang]{zheng2024tracevla}
R.~Zheng, Y.~Liang, S.~Huang, J.~Gao, H.~Daum{\'e}~III, A.~Kolobov, F.~Huang, and J.~Yang.
\newblock Tracevla: Visual trace prompting enhances spatial-temporal awareness for generalist robotic policies.
\newblock \emph{arXiv preprint arXiv:2412.10345}, 2024.

\bibitem[Bjorck et~al.(2025)Bjorck, Castañeda, Cherniadev, Da, Ding, Fan, Fang, Fox, Hu, Huang, Jang, Jiang, Kautz, Kundalia, Lao, Li, Lin, Lin, Liu, Llontop, Magne, Mandlekar, Narayan, Nasiriany, Reed, Tan, Wang, Wang, Wang, Wang, Xiang, Xie, Xu, Xu, Ye, Yu, Zhang, Zhang, Zhao, Zheng, and Zhu]{nvidia2025gr00tn1}
J.~Bjorck, F.~Castañeda, N.~Cherniadev, X.~Da, R.~Ding, L.~J. Fan, Y.~Fang, D.~Fox, F.~Hu, S.~Huang, J.~Jang, Z.~Jiang, J.~Kautz, K.~Kundalia, L.~Lao, Z.~Li, Z.~Lin, K.~Lin, G.~Liu, E.~Llontop, L.~Magne, A.~Mandlekar, A.~Narayan, S.~Nasiriany, S.~Reed, Y.~L. Tan, G.~Wang, Z.~Wang, J.~Wang, Q.~Wang, J.~Xiang, Y.~Xie, Y.~Xu, Z.~Xu, S.~Ye, Z.~Yu, A.~Zhang, H.~Zhang, Y.~Zhao, R.~Zheng, and Y.~Zhu.
\newblock Gr00t n1: An open foundation model for generalist humanoid robots.
\newblock \emph{arXiv preprint arXiv:2503.14734}, 2025.

\bibitem[Deitke et~al.(2024)Deitke, Clark, Lee, Tripathi, Yang, Park, Salehi, Muennighoff, Lo, Soldaini, Lu, Anderson, Bransom, Ehsani, Ngo, Chen, Patel, Yatskar, Callison-Burch, Head, Hendrix, Bastani, VanderBilt, Lambert, Chou, Chheda, Sparks, Skjonsberg, Schmitz, Sarnat, Bischoff, Walsh, Newell, Wolters, Gupta, Zeng, Borchardt, Groeneveld, Dumas, Nam, Lebrecht, Wittlif, Schoenick, Michel, Krishna, Weihs, Smith, Hajishirzi, Girshick, Farhadi, and Kembhavi]{Deitke2024MolmoAP}
M.~Deitke, C.~Clark, S.~Lee, R.~Tripathi, Y.~Yang, J.~S. Park, M.~Salehi, N.~Muennighoff, K.~Lo, L.~Soldaini, J.~Lu, T.~Anderson, E.~Bransom, K.~Ehsani, H.~Ngo, Y.~Chen, A.~Patel, M.~Yatskar, C.~Callison-Burch, A.~Head, R.~Hendrix, F.~Bastani, E.~VanderBilt, N.~Lambert, Y.~Chou, A.~Chheda, J.~Sparks, S.~Skjonsberg, M.~Schmitz, A.~Sarnat, B.~Bischoff, P.~Walsh, C.~Newell, P.~Wolters, T.~Gupta, K.-H. Zeng, J.~Borchardt, D.~Groeneveld, J.~Dumas, C.~Nam, S.~Lebrecht, C.~M. Wittlif, C.~Schoenick, O.~Michel, R.~Krishna, L.~Weihs, N.~A. Smith, H.~Hajishirzi, R.~Girshick, A.~Farhadi, and A.~Kembhavi.
\newblock Molmo and pixmo: Open weights and open data for state-of-the-art multimodal models.
\newblock \emph{arXiv preprint arXiv:2409.17146}, 2024.

\bibitem[Yuan et~al.(2024)Yuan, Duan, Blukis, Pumacay, Krishna, Murali, Mousavian, and Fox]{Yuan2024RoboPointAV}
W.~Yuan, J.~Duan, V.~Blukis, W.~Pumacay, R.~Krishna, A.~Murali, A.~Mousavian, and D.~Fox.
\newblock Robopoint: A vision-language model for spatial affordance prediction for robotics.
\newblock \emph{arXiv preprint arXiv:2406.10721}, 2024.

\bibitem[Shridhar et~al.(2023)Shridhar, Manuelli, and Fox]{shridhar2023perceiver}
M.~Shridhar, L.~Manuelli, and D.~Fox.
\newblock Perceiver-actor: A multi-task transformer for robotic manipulation.
\newblock In \emph{Conference on Robot Learning (CoRL)}, pages 785--799, 2023.

\bibitem[Ma et~al.(2024)Ma, Patidar, Haughton, and James]{ma2024hierarchical}
X.~Ma, S.~Patidar, I.~Haughton, and S.~James.
\newblock Hierarchical diffusion policy for kinematics-aware multi-task robotic manipulation.
\newblock In \emph{Conference on Computer Vision and Pattern Recognition (CVPR)}, pages 18081--18090, 2024.

\bibitem[Yang et~al.(2025)Yang, Cai, Tian, Zeng, and Pang]{yang2025gripper}
Y.~Yang, Z.~Cai, Y.~Tian, J.~Zeng, and J.~Pang.
\newblock Gripper keypose and object pointflow as interfaces for bimanual robotic manipulation.
\newblock \emph{arXiv preprint arXiv:2504.17784}, 2025.

\bibitem[Zhong et~al.(2025)Zhong, Huang, Li, Zhang, Liang, Yang, and Chen]{zhong2025dexgraspvla}
Y.~Zhong, X.~Huang, R.~Li, C.~Zhang, Y.~Liang, Y.~Yang, and Y.~Chen.
\newblock Dexgraspvla: A vision-language-action framework towards general dexterous grasping.
\newblock \emph{arXiv preprint arXiv:2502.20900}, 2025.

\bibitem[Mees et~al.(2022)Mees, Hermann, Rosete-Beas, and Burgard]{mees2022calvin}
O.~Mees, L.~Hermann, E.~Rosete-Beas, and W.~Burgard.
\newblock Calvin: A benchmark for language-conditioned policy learning for long-horizon robot manipulation tasks.
\newblock \emph{Robotics and Automation Letters (RA-L)}, 7\penalty0 (3):\penalty0 7327--7334, 2022.

\bibitem[James et~al.(2020)James, Ma, Rovick~Arrojo, and Davison]{james2019rlbench}
S.~James, Z.~Ma, D.~Rovick~Arrojo, and A.~J. Davison.
\newblock Rlbench: The robot learning benchmark \& learning environment.
\newblock \emph{Robotics and Automation Letters (RA-L)}, 2020.

\bibitem[Niu et~al.(2024)Niu, Sharma, Biamby, Quenum, Bai, Shi, Darrell, and Herzig]{niu2024llarva}
D.~Niu, Y.~Sharma, G.~Biamby, J.~Quenum, Y.~Bai, B.~Shi, T.~Darrell, and R.~Herzig.
\newblock {LLARVA}: Vision-action instruction tuning enhances robot learning.
\newblock In \emph{Conference on Robot Learning (CoRL)}, 2024.

\bibitem[Abeyruwan et~al.(2025)Abeyruwan, Ainslie, Alayrac, Arenas, Armstrong, Balakrishna, Baruch, Bauza, Blokzijl, Bohez, Bousmalis, Brohan, Buschmann, Byravan, Cabi, Caluwaerts, Casarini, Chang, Chen, Chen, Chiang, Choromanski, D'Ambrosio, Dasari, Davchev, Devin, Palo, Ding, Dostmohamed, Driess, Du, Dwibedi, Elabd, Fantacci, Fong, Frey, Fu, Giustina, Gopalakrishnan, Graesser, Hasenclever, Heess, Hernaez, Herzog, Hofer, Humplik, Iscen, Jacob, Jain, Julian, Kalashnikov, Karagozler, Karp, Kew, Kirkland, Kirmani, Kuang, Lampe, Laurens, Leal, Lee, Lee, Liang, Lin, Maddineni, Majumdar, Michaely, Moreno, Neunert, Nori, Parada, Parisotto, Pastor, Pooley, Rao, Reymann, Sadigh, Saliceti, Sanketi, Sermanet, Shah, Sharma, Shea, Shu, Sindhwani, Singh, Soricut, Springenberg, Sterneck, Surdulescu, Tan, Tompson, Vanhoucke, Varley, Vesom, Vezzani, Vinyals, Wahid, Welker, Wohlhart, Xia, Xiao, Xie, Xie, Xu, Xu, Xu, Xu, Yang, Yao, Yaroshenko, Yu, Yuan, Zhang, Zhang, Zhou, and
  Zhou]{geminiroboticsteam2025geminiroboticsbringingai}
S.~Abeyruwan, J.~Ainslie, J.-B. Alayrac, M.~G. Arenas, T.~Armstrong, A.~Balakrishna, R.~Baruch, M.~Bauza, M.~Blokzijl, S.~Bohez, K.~Bousmalis, A.~Brohan, T.~Buschmann, A.~Byravan, S.~Cabi, K.~Caluwaerts, F.~Casarini, O.~Chang, J.~E. Chen, X.~Chen, H.-T.~L. Chiang, K.~Choromanski, D.~D'Ambrosio, S.~Dasari, T.~Davchev, C.~Devin, N.~D. Palo, T.~Ding, A.~Dostmohamed, D.~Driess, Y.~Du, D.~Dwibedi, M.~Elabd, C.~Fantacci, C.~Fong, E.~Frey, C.~Fu, M.~Giustina, K.~Gopalakrishnan, L.~Graesser, L.~Hasenclever, N.~Heess, B.~Hernaez, A.~Herzog, R.~A. Hofer, J.~Humplik, A.~Iscen, M.~G. Jacob, D.~Jain, R.~Julian, D.~Kalashnikov, M.~E. Karagozler, S.~Karp, C.~Kew, J.~Kirkland, S.~Kirmani, Y.~Kuang, T.~Lampe, A.~Laurens, I.~Leal, A.~X. Lee, T.-W.~E. Lee, J.~Liang, Y.~Lin, S.~Maddineni, A.~Majumdar, A.~H. Michaely, R.~Moreno, M.~Neunert, F.~Nori, C.~Parada, E.~Parisotto, P.~Pastor, A.~Pooley, K.~Rao, K.~Reymann, D.~Sadigh, S.~Saliceti, P.~Sanketi, P.~Sermanet, D.~Shah, M.~Sharma, K.~Shea, C.~Shu, V.~Sindhwani, S.~Singh,
  R.~Soricut, J.~T. Springenberg, R.~Sterneck, R.~Surdulescu, J.~Tan, J.~Tompson, V.~Vanhoucke, J.~Varley, G.~Vesom, G.~Vezzani, O.~Vinyals, A.~Wahid, S.~Welker, P.~Wohlhart, F.~Xia, T.~Xiao, A.~Xie, J.~Xie, P.~Xu, S.~Xu, Y.~Xu, Z.~Xu, Y.~Yang, R.~Yao, S.~Yaroshenko, W.~Yu, W.~Yuan, J.~Zhang, T.~Zhang, A.~Zhou, and Y.~Zhou.
\newblock Gemini robotics: Bringing ai into the physical world.
\newblock \emph{arXiv preprint arXiv:2503.20020}, 2025.

\bibitem[Zhen et~al.(2024)Zhen, Qiu, Chen, Yang, Yan, Du, Hong, and Gan]{Zhen20243DVLAA3}
H.~Zhen, X.~Qiu, P.~Chen, J.~Yang, X.~Yan, Y.~Du, Y.~Hong, and C.~Gan.
\newblock 3d-vla: A 3d vision-language-action generative world model.
\newblock \emph{arXiv preprint arXiv:2403.09631}, 2024.

\bibitem[Li et~al.(2024)Li, Hsu, Gu, Pertsch, Mees, Walke, Fu, Lunawat, Sieh, Kirmani, Levine, Wu, Finn, Su, Vuong, and Xiao]{Li2024EvaluatingRR}
X.~Li, K.~Hsu, J.~Gu, K.~Pertsch, O.~Mees, H.~R. Walke, C.~Fu, I.~Lunawat, I.~Sieh, S.~Kirmani, S.~Levine, J.~Wu, C.~Finn, H.~Su, Q.~H. Vuong, and T.~Xiao.
\newblock Evaluating real-world robot manipulation policies in simulation.
\newblock \emph{arXiv preprint arXiv:2405.05941}, 2024.

\bibitem[Wang et~al.(2024)Wang, Zhou, Song, Huang, Shu, and Ma]{Wang2024TowardsTA}
Z.~Wang, Z.~Zhou, J.~Song, Y.~Huang, Z.~Shu, and L.~Ma.
\newblock Towards testing and evaluating vision-language-action models for robotic manipulation: An empirical study.
\newblock \emph{arXiv preprint arXiv:2409.12894}, 2024.

\bibitem[Arai et~al.(2024)Arai, Miwa, Sasaki, Yamaguchi, Watanabe, Aoki, and Yamamoto]{Arai2024CoVLACV}
H.~Arai, K.~Miwa, K.~Sasaki, Y.~Yamaguchi, K.~Watanabe, S.~Aoki, and I.~Yamamoto.
\newblock Covla: Comprehensive vision-language-action dataset for autonomous driving.
\newblock \emph{Winter Conference on Applications of Computer Vision (WACV)}, pages 1933--1943, 2024.

\bibitem[Chen et~al.(2023)Chen, Pourkeshavarz, and Rasouli]{Chen2023CRITERIAAN}
C.~Chen, M.~Pourkeshavarz, and A.~Rasouli.
\newblock Criteria: a new benchmarking paradigm for evaluating trajectory prediction models for autonomous driving.
\newblock \emph{Conference on Robotics and Automation (ICRA)}, pages 8265--8271, 2023.

\bibitem[Duan et~al.(2017)Duan, Andrychowicz, Stadie, Jonathan~Ho, Schneider, Sutskever, Abbeel, and Zaremba]{duan2017oneshot}
Y.~Duan, M.~Andrychowicz, B.~Stadie, O.~Jonathan~Ho, J.~Schneider, I.~Sutskever, P.~Abbeel, and W.~Zaremba.
\newblock One-shot imitation learning.
\newblock In I.~Guyon, U.~V. Luxburg, S.~Bengio, H.~Wallach, R.~Fergus, S.~Vishwanathan, and R.~Garnett, editors, \emph{Advances in Neural Information Processing Systems (NeurIPS)}, volume~30, 2017.

\bibitem[James et~al.(2018)James, Bloesch, and Davison]{james2018task}
S.~James, M.~Bloesch, and A.~J. Davison.
\newblock Task-embedded control networks for few-shot imitation learning.
\newblock \emph{Conference on Robot Learning (CoRL)}, 2018.

\bibitem[Du et~al.(2023)Du, Nair, Sadigh, and Finn]{du2023behavior}
M.~Du, S.~Nair, D.~Sadigh, and C.~Finn.
\newblock Behavior retrieval: Few-shot imitation learning by querying unlabeled datasets.
\newblock In \emph{Proceedings of Robotics: Science and Systems (RSS)}, 2023.

\bibitem[Lin et~al.(2024)Lin, Cui, Xie, Hua, and Sadigh]{lin2024flowretrieval}
L.-H. Lin, Y.~Cui, A.~Xie, T.~Hua, and D.~Sadigh.
\newblock Flowretrieval: Flow-guided data retrieval for few-shot imitation learning.
\newblock In \emph{Conference on Robot Learning (CoRL)}, 2024.

\bibitem[Zhang and Boularias(2024)]{Zhang2024Invariance}
X.~Zhang and A.~Boularias.
\newblock One-shot imitation learning with invariance matching for robotic manipulation.
\newblock In \emph{Proceedings of Robotics: Science and Systems (RSS)}, 2024.

\bibitem[Heppert et~al.(2024)Heppert, Argus, Welschehold, Brox, and Valada]{heppert2024ditto}
N.~Heppert, M.~Argus, T.~Welschehold, T.~Brox, and A.~Valada.
\newblock Ditto: Demonstration imitation by trajectory transformation.
\newblock In \emph{Conference on Intelligent Robots and Systems (IROS)}, 2024.

\bibitem[Di~Palo and Johns(2024)]{dipalo2024kat}
N.~Di~Palo and E.~Johns.
\newblock Keypoint action tokens enable in-context imitation learning in robotics.
\newblock In \emph{Proceedings of Robotics: Science and Systems (RSS)}, 2024.

\bibitem[Fu et~al.(2024)Fu, Huang, Datta, Chen, Panitch, Liu, Li, and Goldberg]{fu2024context}
L.~Fu, H.~Huang, G.~Datta, L.~Y. Chen, W.~C.-H. Panitch, F.~Liu, H.~Li, and K.~Goldberg.
\newblock In-context imitation learning via next-token prediction.
\newblock \emph{arXiv preprint arXiv:2408.15980}, 2024.

\bibitem[Steiner et~al.(2024)Steiner, Pinto, Tschannen, Keysers, Wang, Bitton, Gritsenko, Minderer, Sherbondy, Long, Qin, Ingle, Bugliarello, Kazemzadeh, Mesnard, Alabdulmohsin, Beyer, and Zhai]{Steiner2024PaliGemma2A}
A.~Steiner, A.~S. Pinto, M.~Tschannen, D.~Keysers, X.~Wang, Y.~Bitton, A.~Gritsenko, M.~Minderer, A.~Sherbondy, S.~Long, S.~Qin, R.~R. Ingle, E.~Bugliarello, S.~Kazemzadeh, T.~Mesnard, I.~M. Alabdulmohsin, L.~Beyer, and X.-Q. Zhai.
\newblock Paligemma 2: A family of versatile vlms for transfer.
\newblock \emph{arXiv preprint arXiv:2412.03555}, 2024.

\bibitem[Brohan et~al.(2022)Brohan, Brown, Carbajal, Chebotar, Dabis, Finn, Gopalakrishnan, Hausman, Herzog, Hsu, Ibarz, Ichter, Irpan, Jackson, Jesmonth, Joshi, Julian, Kalashnikov, Kuang, Leal, Lee, Levine, Lu, Malla, Manjunath, Mordatch, Nachum, Parada, Peralta, Perez, Pertsch, Quiambao, Rao, Ryoo, Salazar, Sanketi, Sayed, Singh, Sontakke, Stone, Tan, Tran, Vanhoucke, Vega, Vuong, Xia, Xiao, Xu, Xu, Yu, and Zitkovich]{Brohan2022RT1RT}
A.~Brohan, N.~Brown, J.~Carbajal, Y.~Chebotar, J.~Dabis, C.~Finn, K.~Gopalakrishnan, K.~Hausman, A.~Herzog, J.~Hsu, J.~Ibarz, B.~Ichter, A.~Irpan, T.~Jackson, S.~Jesmonth, N.~J. Joshi, R.~C. Julian, D.~Kalashnikov, Y.~Kuang, I.~Leal, K.-H. Lee, S.~Levine, Y.~Lu, U.~Malla, D.~Manjunath, I.~Mordatch, O.~Nachum, C.~Parada, J.~Peralta, E.~Perez, K.~Pertsch, J.~Quiambao, K.~Rao, M.~S. Ryoo, G.~Salazar, P.~R. Sanketi, K.~Sayed, J.~Singh, S.~A. Sontakke, A.~Stone, C.~Tan, H.~Tran, V.~Vanhoucke, S.~Vega, Q.~H. Vuong, F.~Xia, T.~Xiao, P.~Xu, S.~Xu, T.~Yu, and B.~Zitkovich.
\newblock Rt-1: Robotics transformer for real-world control at scale.
\newblock \emph{arXiv preprint arXiv:2212.06817}, 2022.

\bibitem[Ning et~al.(2023)Ning, Li, Zhang, Geng, Dai, He, and Hu]{Ning2023AllIT}
J.~Ning, C.~Li, Z.~Zhang, Z.~Geng, Q.~Dai, K.~He, and H.~Hu.
\newblock All in tokens: Unifying output space of visual tasks via soft token.
\newblock \emph{Conference on Computer Vision (ICCV)}, pages 19843--19853, 2023.

\bibitem[Deitke et~al.(2023)Deitke, Schwenk, Salvador, Weihs, Michel, VanderBilt, Schmidt, Ehsani, Kembhavi, and Farhadi]{deitke2023objaverse}
M.~Deitke, D.~Schwenk, J.~Salvador, L.~Weihs, O.~Michel, E.~VanderBilt, L.~Schmidt, K.~Ehsani, A.~Kembhavi, and A.~Farhadi.
\newblock Objaverse: A universe of annotated 3d objects.
\newblock In \emph{Conference on Computer Vision and Pattern Recognition (CVPR)}, pages 13142--13153, 2023.

\bibitem[Johnson et~al.(2016)Johnson, Hariharan, van~der Maaten, Fei-Fei, Zitnick, and Girshick]{Johnson2016CLEVRAD}
J.~Johnson, B.~Hariharan, L.~van~der Maaten, L.~Fei-Fei, C.~L. Zitnick, and R.~B. Girshick.
\newblock Clevr: A diagnostic dataset for compositional language and elementary visual reasoning.
\newblock \emph{Conference on Computer Vision and Pattern Recognition (CVPR)}, pages 1988--1997, 2016.

\bibitem[Ehsani et~al.(2024)Ehsani, Gupta, Hendrix, Salvador, Weihs, Zeng, Singh, Kim, Han, Herrasti, et~al.]{ehsani2024spoc}
K.~Ehsani, T.~Gupta, R.~Hendrix, J.~Salvador, L.~Weihs, K.-H. Zeng, K.~P. Singh, Y.~Kim, W.~Han, A.~Herrasti, et~al.
\newblock Spoc: Imitating shortest paths in simulation enables effective navigation and manipulation in the real world.
\newblock In \emph{Conference on Computer Vision and Pattern Recognition (CVPR)}, pages 16238--16250, 2024.

\bibitem[Greff et~al.(2022)Greff, Belletti, Beyer, Doersch, Du, Duckworth, Fleet, Gnanapragasam, Golemo, Herrmann, Kipf, Kundu, Lagun, Laradji, Liu, Meyer, Miao, Nowrouzezahrai, Oztireli, Pot, Radwan, Rebain, Sabour, Sajjadi, Sela, Sitzmann, Stone, Sun, Vora, Wang, Wu, Yi, Zhong, and Tagliasacchi]{Greff2022KubricAS}
K.~Greff, F.~Belletti, L.~Beyer, C.~Doersch, Y.~Du, D.~Duckworth, D.~J. Fleet, D.~Gnanapragasam, F.~Golemo, C.~Herrmann, T.~Kipf, A.~Kundu, D.~Lagun, I.~H. Laradji, H.-T. Liu, H.~Meyer, Y.~Miao, D.~Nowrouzezahrai, C.~Oztireli, E.~Pot, N.~Radwan, D.~Rebain, S.~Sabour, M.~S.~M. Sajjadi, M.~Sela, V.~Sitzmann, A.~Stone, D.~Sun, S.~Vora, Z.~Wang, T.~Wu, K.~M. Yi, F.~Zhong, and A.~Tagliasacchi.
\newblock Kubric: A scalable dataset generator.
\newblock \emph{Conference on Computer Vision and Pattern Recognition (CVPR)}, pages 3739--3751, 2022.

\bibitem[Quattoni and Torralba(2009)]{Quattoni2009RecognizingIS}
A.~Quattoni and A.~Torralba.
\newblock Recognizing indoor scenes.
\newblock \emph{Conference on Computer Vision and Pattern Recognition (CVPR)}, pages 413--420, 2009.

\bibitem[Fang et~al.(2022)Fang, Wang, Fang, Gou, Liu, Yan, Liu, Xie, and Lu]{Fang2022AnyGraspRA}
H.~Fang, C.~Wang, H.~Fang, M.~Gou, J.~Liu, H.~Yan, W.~Liu, Y.~Xie, and C.~Lu.
\newblock Anygrasp: Robust and efficient grasp perception in spatial and temporal domains.
\newblock \emph{Transactions on Robotics (T-RO)}, 39:\penalty0 3929--3945, 2022.

\bibitem[Zhou et~al.(2024)Zhou, Blessing, Li, Celik, Jia, Neumann, and Lioutikov]{Zhou2024VariationalDO}
H.~Zhou, D.~Blessing, G.~Li, O.~Celik, X.~Jia, G.~Neumann, and R.~Lioutikov.
\newblock Variational distillation of diffusion policies into mixture of experts.
\newblock \emph{arXiv preprint arXiv:2406.12538}, 2024.

\bibitem[Tschannen et~al.(2025)Tschannen, Gritsenko, Wang, Naeem, Alabdulmohsin, Parthasarathy, Evans, Beyer, Xia, Mustafa, H'enaff, Harmsen, Steiner, and Zhai]{Tschannen2025SigLIP2M}
M.~Tschannen, A.~Gritsenko, X.~Wang, M.~F. Naeem, I.~M. Alabdulmohsin, N.~Parthasarathy, T.~Evans, L.~Beyer, Y.~Xia, B.~Mustafa, O.~H'enaff, J.~Harmsen, A.~Steiner, and X.-Q. Zhai.
\newblock Siglip 2: Multilingual vision-language encoders with improved semantic understanding, localization, and dense features.
\newblock \emph{arXiv preprint arXiv:2502.14786}, 2025.

\bibitem[Lin et~al.(2014)Lin, Maire, Belongie, Hays, Perona, Ramanan, Doll{\'a}r, and Zitnick]{lin2014microsoft}
T.-Y. Lin, M.~Maire, S.~Belongie, J.~Hays, P.~Perona, D.~Ramanan, P.~Doll{\'a}r, and C.~L. Zitnick.
\newblock Microsoft coco: Common objects in context.
\newblock In \emph{Computer vision--ECCV 2014: 13th European conference, zurich, Switzerland, September 6-12, 2014, proceedings, part v 13}, pages 740--755. Springer, 2014.

\end{thebibliography}

\appendix
\clearpage


{\Large{\bf Supplementary Material}}\\

\section{Pose and Trajectory L1 Metric}
\label{app:sec:lonemetric}
We define an L1-based metric for pose error that combines the positional and rotational components in a single scalar. The position error is computed as the L1 norm of the difference between predicted and ground truth translation vectors. The rotation error is measured in degrees, and we normalize both terms using the equivalence of \textbf{1\,cm = 1°}. The overall pose L1 error is given by:

\begin{equation}
\text{mean taj. L1} = \frac{1}{N} \sum_{i=1}^{N} \left( \| \mathbf{t}_i - \hat{\mathbf{t}}_i \|_1 + \angle\left( \hat{R}_i^{-1} R_i \right) \right)
\end{equation}

\noindent where:
\begin{itemize}
  \item $N$ is the number of poses in the trajectory.
  \item $\mathbf{t}_i,\hat{\mathbf{t}}_i \in \mathbb{R}^3$ are the predicted and ground truth translation vectors for sample $i$.
  \item $R_i, \hat{R}_i \in \mathrm{SO}(3)$ are the predicted and ground truth rotation matrices.
  \item $\angle(\cdot)$ denotes the angle (in degrees) of the relative rotation.
\end{itemize}

\section{Simulation Setup}
\label{app:sec:simulation}
We make use of the Maniskill3 simulation environment~\cite{Tao2024ManiSkill3GP}. For each interaction, we predict two waypoints corresponding to the grasping and gripper opening position. These are converted into a trajectory by adding a raising and destination alignment movement. These are then executed using an inverse kinematics planning system. The reward is computed by comparing the L2 norm between the object's current and goal poses. This quantity is mapped to the unit range by normalizing it with the initial distance between the initial pose and the goal and computing 1 - this quantity. The resulting rewards are clipped to the unit range. Success rates are computed using a reward threshold of 0.75. 

\begin{equation}
\text{reward} = \text{clamp}\left(1 - \frac{\lVert \mathbf{p}_A - \mathbf{p}_{\text{goal}} \rVert}{\lVert \mathbf{p}_A^{\text{init}} - \mathbf{p}_{\text{goal}} \rVert},\ [0,\ 1]\right)
\end{equation}
\noindent
where:
\begin{itemize}
    \item \( \mathbf{p}_A \in \mathbb{R}^3 \) is the current position of object,
    \item \( \mathbf{p}_A^{\text{init}} \in \mathbb{R}^3 \) is the initial position of the object at the start of the episode,
    \item \( \mathbf{p}_{\text{goal}} \in \mathbb{R}^3 \) is the target goal position for object
\end{itemize}

\section{Objaverse Asset Curation}

\label{sec:objaverse_curation}
Objaverse assets were created by a mix of artists, the sizes of the objects are not scaled canonically, and the provided category label is not fine-grained enough to generate descriptions. We follow the following procedure to create a large dataset of diverse high-quality models with good, concise text descriptions. We start by subsampling the dataset of 1M shapes to 600k. Then, to obtain the relevant meta-information of model scale as well as text description, we start by using GPT-4V to produce long-form descriptions including a guess of the object's dimensions, based on several rendered perspectives. We do an intermediate filtering step where each object is filtered by size: a) only objects with all side lengths between 0.01 m and 0.20 m are retained, and b) objects with a side length ratio exceeding 5 (i.e., too elongated) are excluded. After filtering, objects are rescaled such that the smallest side in the x–y plane is at most 0.07 m, ensuring compatibility with the gripper size (0.08 m), including margin.

To further obtain concise descriptions, we use GPT-4 to summarize the long-form descriptions. In the final verification step, the short form descriptions were evaluated using SigLIP~\citep{Tschannen2025SigLIP2M}. Specifically, we compare the embeddings of the images against the short descriptions. Further, we use SigLIP to evaluate the alignment of the short descriptions with the captions “grayscale image” and "cartoon low-poly model”. Eventually, we only select objects for our asset set where the distance between SigLIP embeddings passes a threshold, resulting in a final object set of around 7k models.

\section{Simulation and Real Experiments}
\label{app:sec:datasets}

\begin{figure}[ht]
    \centering
    \begin{subfigure}[t]{0.45\textwidth}
        \centering
        \includegraphics[width=\linewidth]{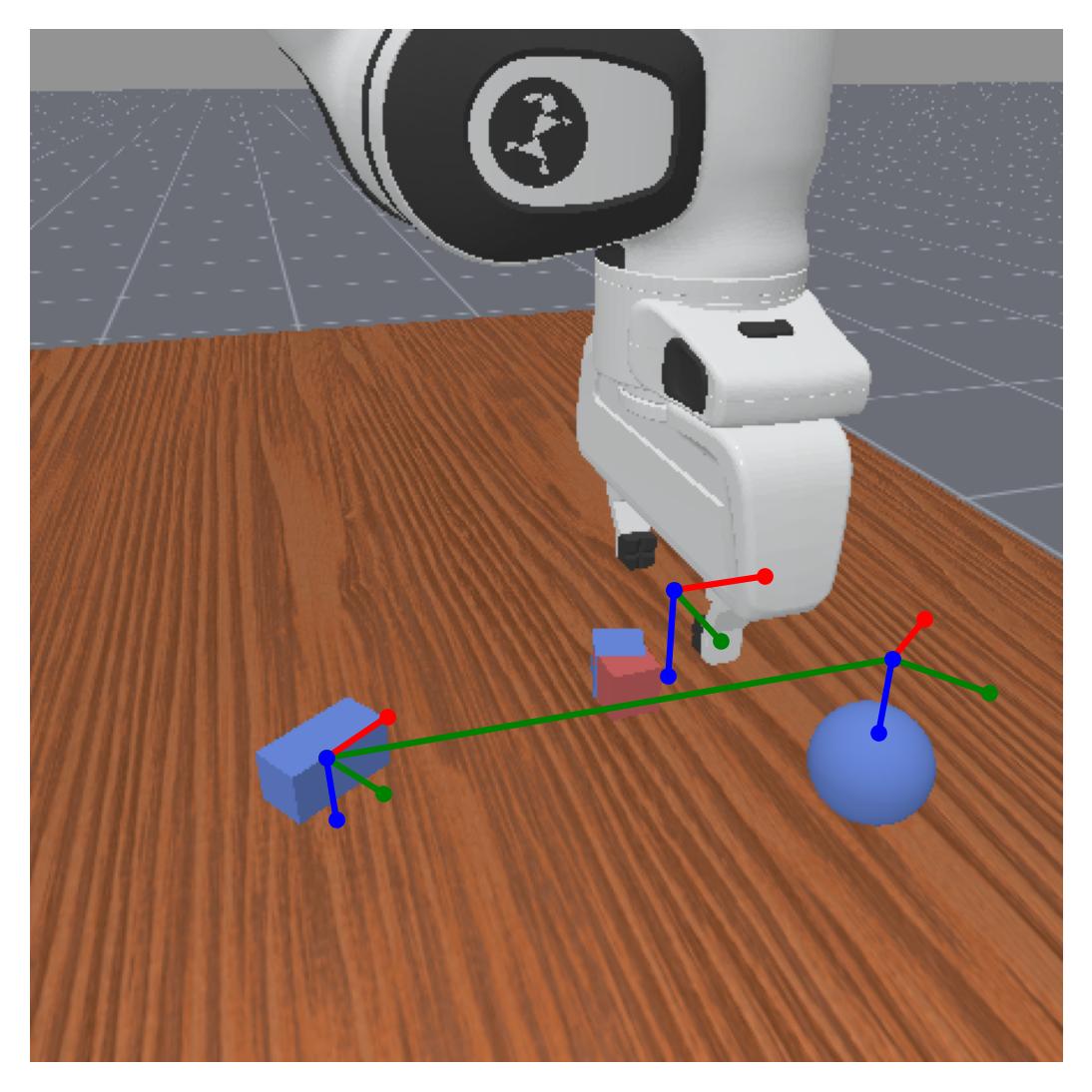}
        \includegraphics[width=\linewidth]{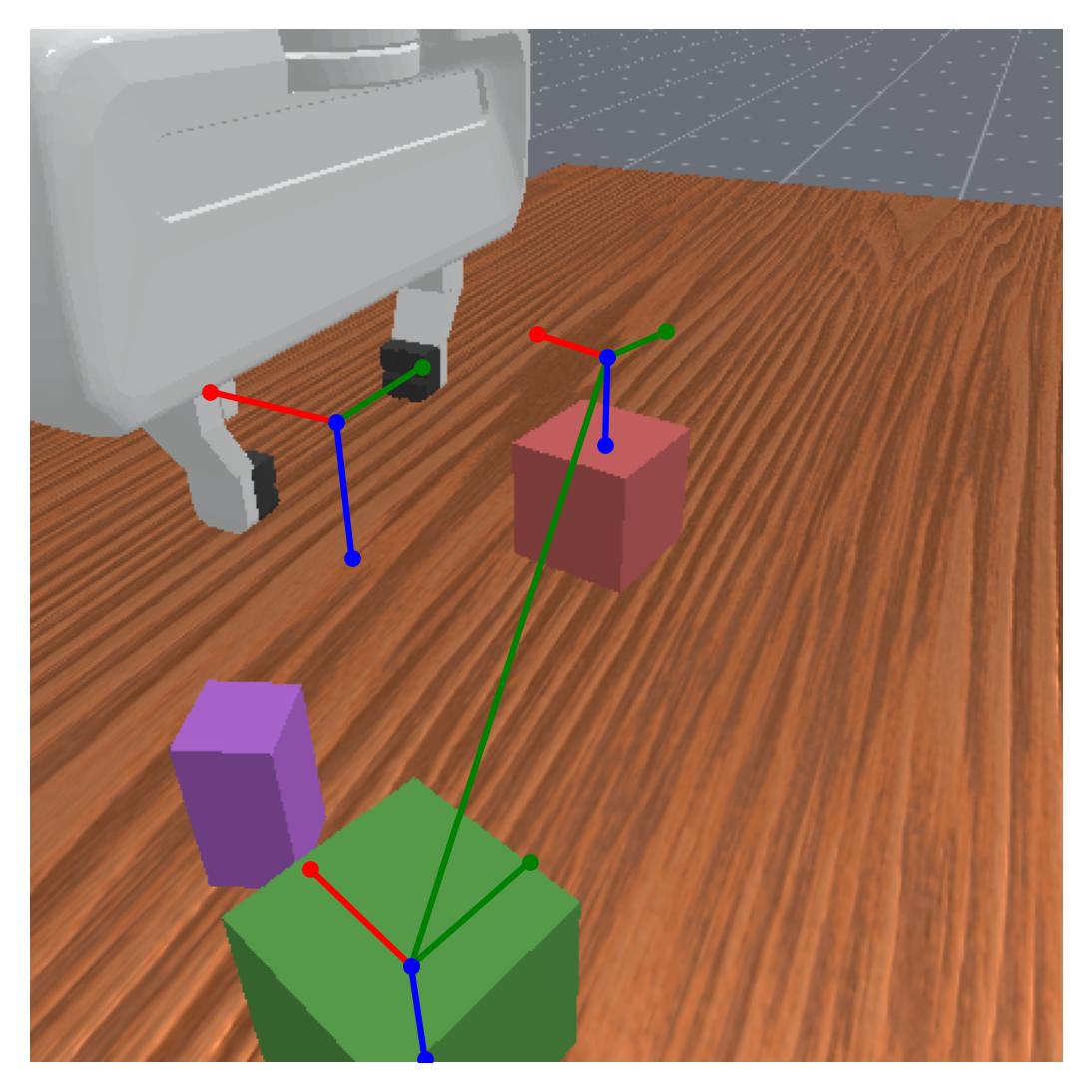}
        \caption{CLEVR assets}
    \end{subfigure}
    \hfill
    \begin{subfigure}[t]{0.45\textwidth}
        \centering
        \includegraphics[width=\linewidth]{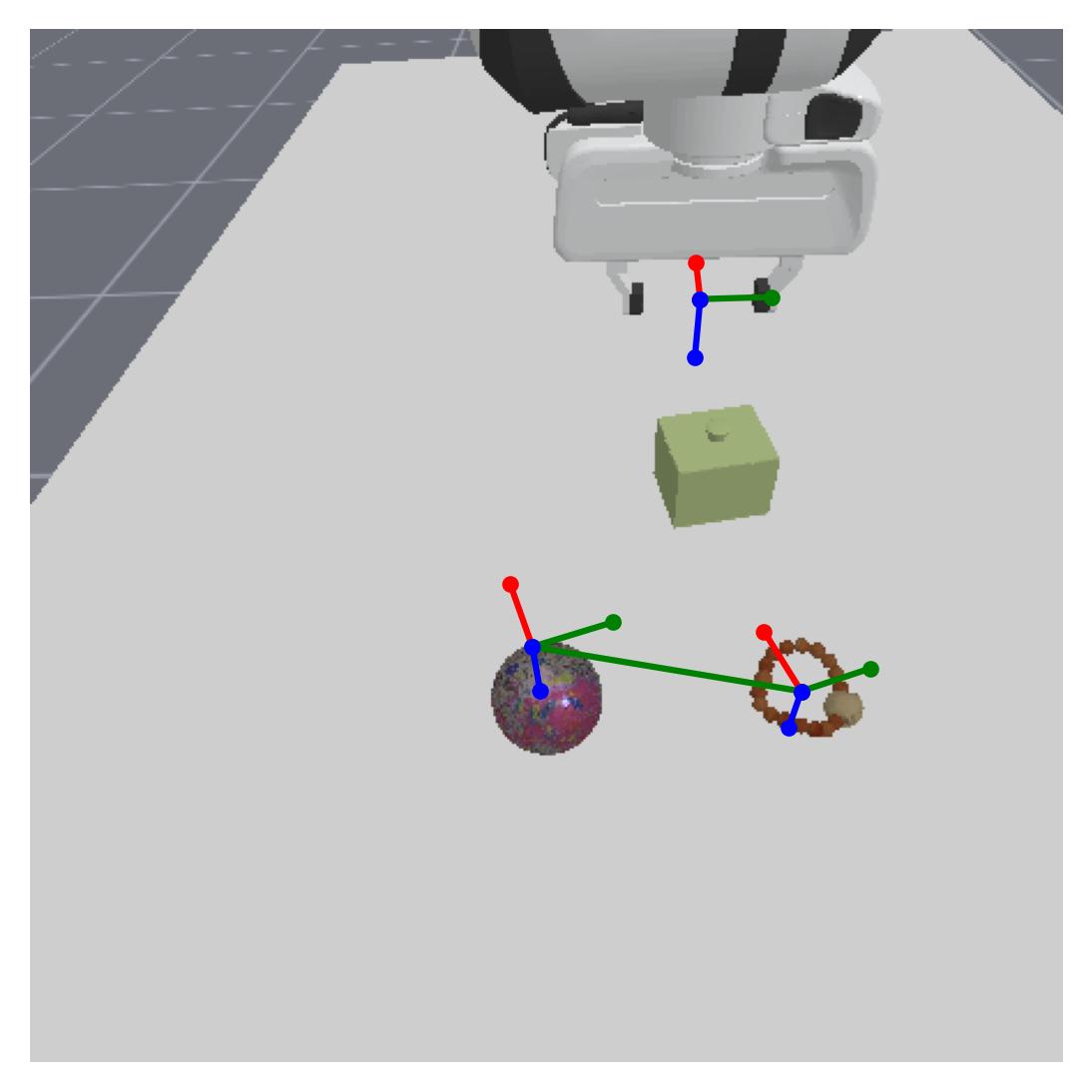}
        \includegraphics[width=\linewidth]{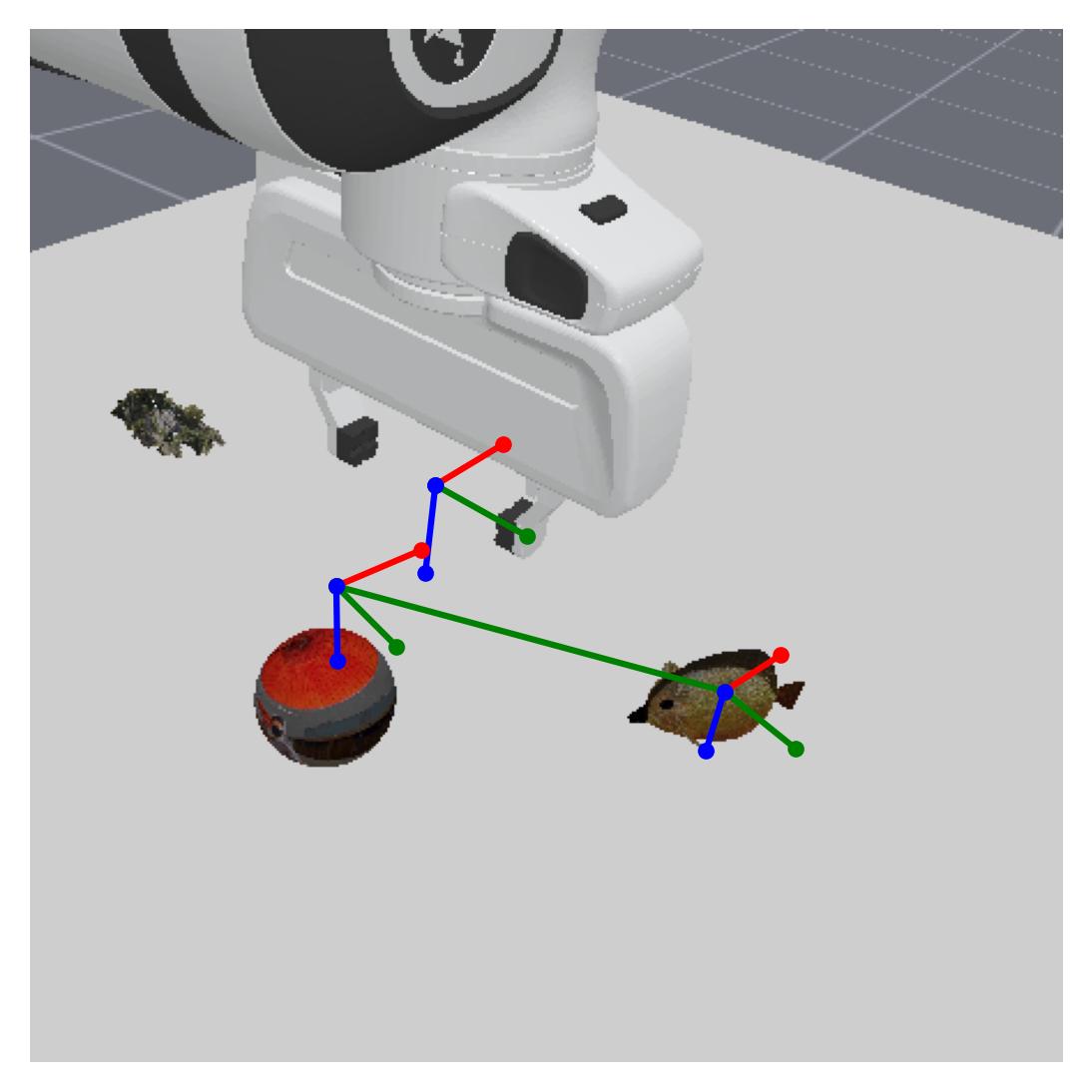}
        \caption{Objaverse assets}
    \end{subfigure}
    \caption{\textbf{Qualitative examples of our training data.} Our training data consists of different 3D objects placed into the scene. We distinguish between 2 different object groups - CLEVR-like assets and Objaverse assets.}
    \label{fig:dataset_qualitative_examples}
\end{figure}

We use different versions of the training and evaluation datasets in our pipeline. Training datasets are curated using ManiSkill3 simulation environment~\cite{Tao2024ManiSkill3GP}, where the objects in the scene come from either CLEVR~\cite{Johnson2016CLEVRAD} or Objaverse~\cite{deitke2023objaverse} datasets. We further introduce two difficulty levels of the training datasets \textit{\textbf{CLEVR/Objaverse-easy}} and \textit{\textbf{CLEVR/Objaverse-hard}}, where the difference is in the scene and camera field-of-view randomization between them. \cref{fig:dataset_qualitative_examples} shows examples from the harder version of the training datasets with both CLEVR and Objaverse assets.

We evaluated our models on four different application domains - simulated data, real data, simulations, and real robot setups. Evaluations on simulated data and simulations were performed in the same setup as the training data was curated, but with new environments and scenes. 

For the real data evaluations, we used different subsets of the DROID dataset~\cite{Khazatsky2024DROIDAL}. This data first needs to be filtered as the quality of the extrinsic calibration is very variable, something that has been noted and systemically addressed in recent work \href{https://medium.com/@zubair_irshad/scaling-up-automatic-camera-calibration-for-droid-dataset-4ddfc45361d3}{here.}

For our small-scale evaluation sets, we manually filtered the data using the projection of the end-effector position into the image. For our testing purposes, after manual filtering based on calibration, we selected text prompts containing the word ``block" and took 160 episodes without background clutter, which made our \textit{\textbf{DROID-hard}} subset. This subset contains images of blocks as well as a few visually confounding items. The presence of these leads to false predictions and makes it suitable to evaluate predictions of multiple trajectories. To make predictions without confounding objects, we created \textit{\textbf{DROID-easy}} subset, which blurred out the confounding objects, see \cref{fig:one_shot_visuals} for example.

For real robot evaluations, we used a Franka Panda robotic arm. The real robot was run using a ROS setup, the inverse kinematics planning was done using \texttt{bio\_ik} package.
We used the version of our network without depth inputs, as we did not implement any depth augmentation, to compensate for the missing depth we projected the grasp point onto the closes valid depth value along the position ray.

\section{Training Details}
\label{app:sec:trainig_details}

In \cref{tab:hyperparameters}, we provide an overview of the hyperparameters that we use in our experiments. We leverage the Hugging Face library and the pretrained PaliGemma2~\cite{Steiner2024PaliGemma2A} model for fine-tuning.  

\begin{table}[ht]
    \footnotesize
    \centering
    \begin{tabular}{ll}
        \toprule
        \textbf{Hyperparameter} & \textbf{Value} \\
        \midrule
        Learning rate & 3e-5 \\
        Learning rate scheduler & cosine \\
        Warmup ratio & 0.05 \\
        Optimizer & Adafactor \\
        Batch size train & 32 \\
        Training epochs & 1\\
        Training iterations & 4687 (150k samples) \\
        Trainable layers & Self-attention layers only \\
        \bottomrule
    \end{tabular}
    \caption{Training hyperparameters used throughout in our experiments.}
    \label{tab:hyperparameters}
\end{table}

\section{One-shot Imitation from Demonstrations}
\label{app:sec:one_shot_details}

Here we provide more information about extending our system to support few-shot imitation from demonstrations. We provide examples of the used prompts and visualize the predictions from simulations and real-data evaluations. 

\begin{figure}[ht]
    \centering
    \begin{tabular}{@{}c@{}c@{}cc@{}}
    \multicolumn{3}{c}{Inputs} & Output \\
    \cline{1-3} \\
    \begin{minipage}[t]{0.2\textwidth}
        \centering
        \includegraphics[width=\linewidth]{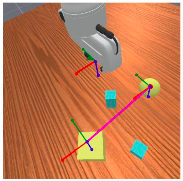}
    \end{minipage}
    &
    \hfill
    \begin{minipage}[t]{0.29\textwidth}
        \vspace{-5em}
        \tiny \texttt{<loc0243>}\texttt{<loc0423>}\texttt{<loc0751>} \texttt{<seg063>}\texttt{<seg079>}\texttt{<seg112>} \texttt{<loc0403>}\texttt{<loc0241>}\texttt{<loc0732>} \texttt{<seg063>}\texttt{<seg079>}\texttt{<seg112>}
    \end{minipage}
    \hfill
    &
    \begin{minipage}[t]{0.2\textwidth}
        \centering
        \includegraphics[trim=0 1pt 0 1pt, clip, width=\linewidth]{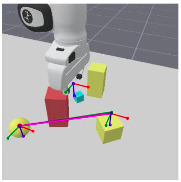}
    \end{minipage}
    \hfill
    &
    \begin{minipage}[t]{0.29\textwidth}
        \vspace{-5em}
        \tiny
        \texttt{<loc0354>}\texttt{<loc0050>}\texttt{<loc0772>} \texttt{<seg045>}\texttt{<seg067>}\texttt{<seg071>} \texttt{<loc0314>}\texttt{<loc0317>}\texttt{<loc0768>} \texttt{<seg045>}\texttt{<seg067>}\texttt{<seg071>}
    \end{minipage}

    \\
    
    \texttt{<demo img>} & \texttt{<demo traj.>} & \texttt{<live img>} & \texttt{<est. traj.>}\\
    \end{tabular}

    \caption{\textbf{One-shot prompt example.} Prompts consist of a demonstration image-trajectory path and a live image for which the model should predict the trajectory of the task represented with demonstration data. No language description is provided, yet the model is able to imitate the task. The given prompt represents the task, "move large yellow sphere onto large yellow cube". The pink line in the first image presents the demonstration trajectory, while the pink line in the second image presents the predicted trajectory. Ground truth of the live image trajectory is presented with a green line, but it is not visible due to overlapping. Robot state inputs not included for brevity.}
    \label{fig:one_shot_prompt}
\end{figure}

Our imitation extension requires a specific prompt format which follows the template of \texttt{<demo img>} + \texttt{<demo robot state>} + \texttt{<demo trajectory>} + \texttt{<live img>} + \texttt{<live robot state>} $\rightarrow$ \texttt{<estimated trajectory>}. We further provide the information of the robot state after the images. 
An example of one prompt is shown in \cref{fig:one_shot_prompt}, note that no explicit text description is given, only the tokens of the demonstration trajectory.

\begin{figure}[hb]
    \centering
    \begin{subfigure}[t]{0.3\textwidth}
        \centering
        \includegraphics[width=\linewidth]{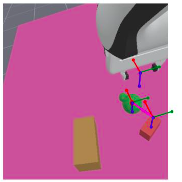}
        \includegraphics[width=\linewidth]{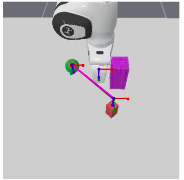}
        \caption{\textbf{CLEVR-hard:} move large green sphere onto small red box}
    \end{subfigure}
    \hfill
    \begin{subfigure}[t]{0.3\textwidth}
        \centering
        \includegraphics[trim=0 0 2pt 2pt, clip, width=\linewidth]{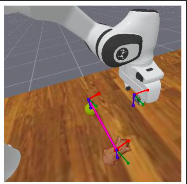}
        \includegraphics[trim=0 0 2pt 0, clip, width=\linewidth]{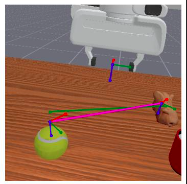}
        \caption{\textbf{Objaverse-easy:} move terracotta animal onto yellow ball}
    \end{subfigure}
    \hfill 
    \begin{subfigure}[t]{0.3\textwidth}
        \centering
        \includegraphics[width=\linewidth]{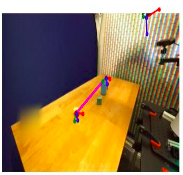}
        \includegraphics[width=\linewidth]{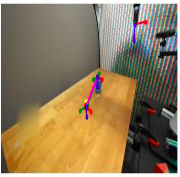}
        \caption{\textbf{DROID-easy:} put the yellow block inside the light blue cup}
    \end{subfigure}

    \caption{\textbf{Demonstration and live image with predictions.} Examples of demonstration image in the first row with visualized ground truth trajectory (pink line) and live image with both ground truth (green line) and predicted trajectory (pink line) in the second row for three different datasets - simulation data matching the training distribution, the Objaverse dataset, and DROID easy. Our imitation model performs well on all three application domains, showing good generalization to imitation with completely unseen objects in the Objaverse dataset.}
    \label{fig:one_shot_visuals}
\end{figure}

To create demonstration - live image pairs, we implemented a look-up table in the dataset. We sample demonstration–live image pairs such that each unique combination is used only once and never repeated during training. For the evaluation, we are using a hold-out validation dataset with the same distribution as the training, but in new environments. We again repeat the sampling process and conduct an evaluation over 10k pre-sampled combinations. When evaluating in simulation, we use a hold-out validation dataset to fetch a demonstration pair that corresponds to the given simulation task. 
\cref{fig:one_shot_visuals} shows predictions of the imitation model trained on CLEVR-hard dataset version. Our imitation model generalizes well to different application domains, especially to the Objaverse dataset, where the imitation is performed with completely unseen objects. 

\section{Input Image Cropping}
\label{sec:cropping_details}

We evaluate the effect of different crop strategies on performance by choosing different crop centers and applying zero padding, as shown in \cref{fig:effect_of_crop_size_block_v3}. 
When cropping, the region of interest around the task-relevant objects is enlarged before processing. 
The center is set to: 1) image center, 2) start object, 3) middle point between start and end objects. The crop of size $w\times w$ is taken without zero-padding (valid mode) or with zero-padding (non-valid mode) and resized to model image resolution $224\times 224$. Absence of padding doesn't preserve aspect ratio.
However, we emphasize that solving object scale sensitivity is not the primary focus of this work, and we leave more general solutions (e.g., multiscale feature extraction or higher-resolution processing) to future research.

\section{Beam-Search-NMS  Implementation}
\label{sec:app-beam-search-NMS}
We propose a variant of beam-search that also does non-maximum suppression over a span of spatially contiguous tokens. This is done in the following manner: a Point $x$ is a local maximum if $p(x) \geq p(x')\ \forall x' \in [x-w,x+w]$. With a noisy distribution, $w$ should be larger; however, with too large $w$, we will suppress all maxima except the global one. Thus, we find that $w=100$ was sufficiently large. Our decoding procedure is beam search with $ n=3$. After processing the coordinate or angle distribution with NMS and suppressing all non-maxima by setting the token log-probability to $-\infty$, the next top $ n$ tokens are selected. 

We compare our beam search-NMS with other decoding strategies using trajectory mean L1 error (\cref{tab:trajectory_errors}). Additionally, we plot the beam score (log-probability) vs. the mean L1 error of the trajectory. Both standard beam search and beam search-NMS show correlation between beam log-probability and error (\cref{fig:beam_seatch_err_correlation}), which enables us to compute a precision-recall curve in object detection style by replacing IoU with L1 measure (\cref{fig:precision_recall_curve}). Our NMS approach has an mAP of 0.31 on original size DROID-hard images, while standard beam search has an mAP of 0.11.

\section{mAP Calculation}
\label{app:sec:additional_experimental}

\begin{figure}[hb]
    \centering
    \begin{minipage}{.5\textwidth}
      \centering
       \includegraphics[width=.9\columnwidth, trim=0cm 0cm 0cm .8cm, clip]{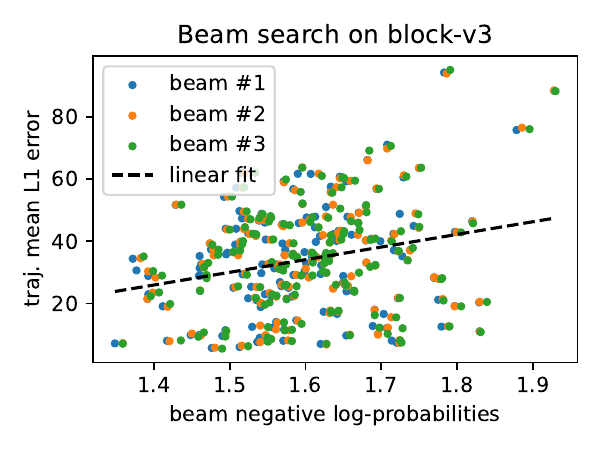}
      \caption*{a) Standard beam search}
    \end{minipage}\hfill
    \begin{minipage}{.5\textwidth}
      \centering
       \includegraphics[width=.9\columnwidth, trim=0cm 0cm 0cm .8cm, clip]{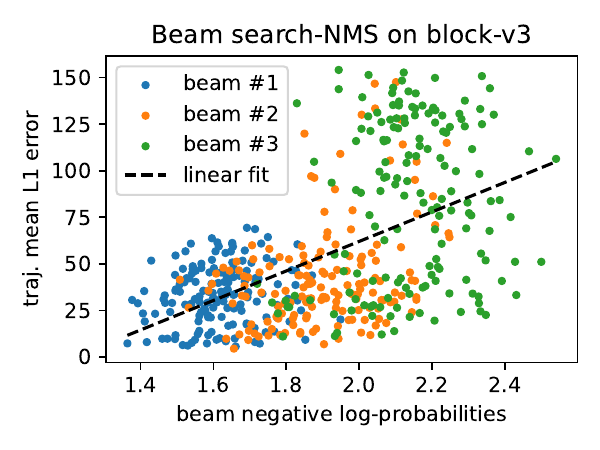}
      \caption*{b) Beam Search-NMS (ours)}
    \end{minipage}
    \caption{\textbf{Beam log-pobs correlate with L1 error.} Higher negative log-prob means lower confidence. Existence of correlation allows us to use log-probs as confidences. Results show an exploration-exploitation tradeoff. Standard beam search generates low error samples, these predictions are not very diverse, see a). Our approach generates more diverse predictions, allowing it to explore other possible trajectories. The spearman rank coefficients are 0.17 and 0.49 respectively.}
    \label{fig:beam_seatch_err_correlation}
\end{figure}

\begin{figure}[ht]
    \centering
    \begin{subfigure}[t]{.48\textwidth}
      \centering
        \includegraphics[width=1.0\columnwidth]{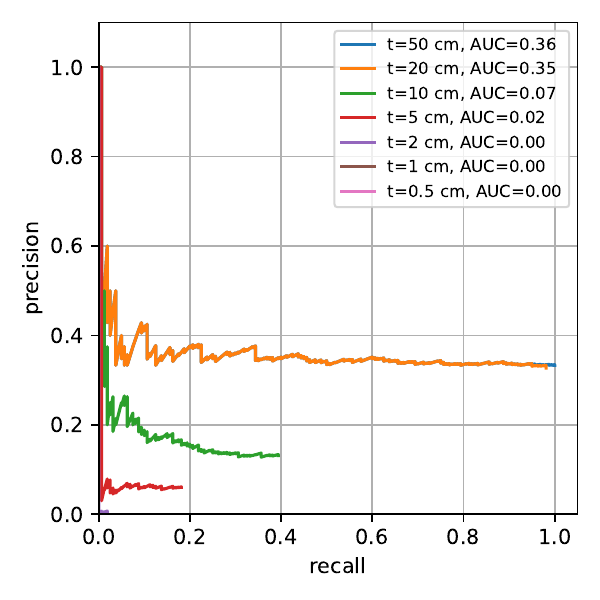}
        \caption{Standard beam search, no crop, mAP=0.11}
    \end{subfigure}
    \hfill
    \begin{subfigure}[t]{.48\textwidth}
      \centering
        \includegraphics[width=1.0\columnwidth]{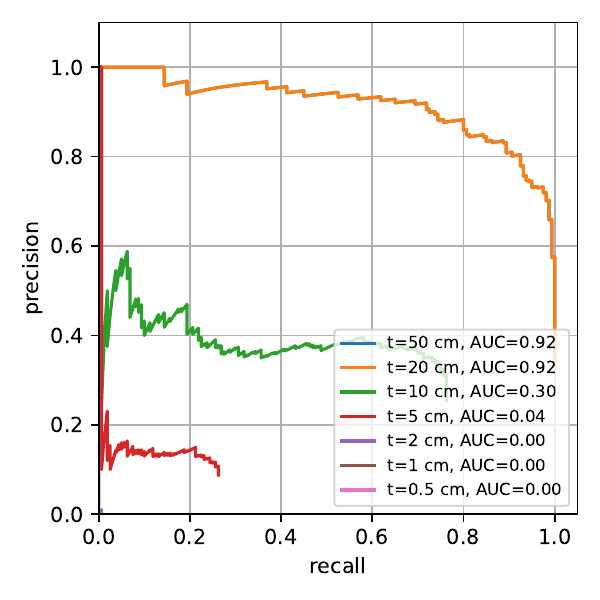}
        \caption{Beam search-NMS (ours), no crop, mAP=0.31}
    \end{subfigure}
    \\
    \begin{subfigure}[t]{.48\textwidth}
      \centering
        \includegraphics[width=1.0\columnwidth]{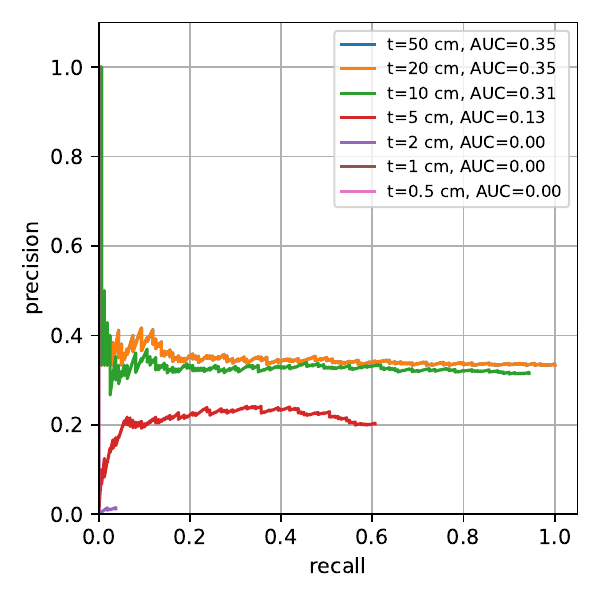}
        \caption{Standard beam search, crop size 700, mAP=0.16}
    \end{subfigure}
    \hfill
    \begin{subfigure}[t]{.48\textwidth}
      \centering
        \includegraphics[width=1.0\columnwidth]{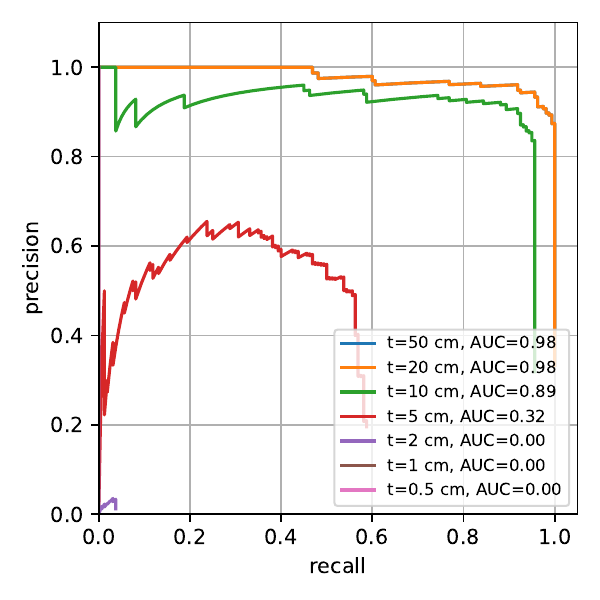}
        \caption{Beam search-NMS (ours), crop size 700, mAP=0.45}
    \end{subfigure}
    \caption{Precision-recall curves at different error thresholds. Upper row -- no crop, lower row -- crop size 700 with padding. Object sizes are around 10 cm, thus we suggest $\text{mAP}_{[0.5\,50]}$, meaning the mAP at thresholds of [.5, 1, 2, 5, 10, 20, 50]~cm, with 1cm = 10 degrees. 
    }
    \label{fig:precision_recall_curve}
\end{figure}

To evaluate the generation of multiple trajectories, we adapt the mean Average Precision (mAP) metric from the COCO object detection challenge ~\cite{lin2014microsoft}:

\begin{enumerate}
    \item Instead of bounding boxes, we compare trajectories.
    \item For each episode, we have one ground-truth trajectory and several predictions, for the confidence of predictions, we use the log-probability of the beam, see \cref{fig:beam_seatch_err_correlation}.
    \item We replace intersection over union (IoU) with the mean L1 error over keyposes, a prediction is considered a false positive if the L1 error is larger than a threshold. See \cref{fig:precision_recall_curve}.
\end{enumerate}

In this metric, we don't average of classes, only different values of L1 thresholds. 
Using this metric, we compared our beam search-NMS with standard beam search, as the latter is the second-best variant on the DROID-hard dataset \cref{tab:trajectory_errors}. The mAP and precision-recall curves are shown in \cref{fig:precision_recall_curve}.

\end{document}